\documentclass[preprint,3p,10pt]{elsarticle}
\usepackage{graphicx, subcaption}
\usepackage{picinpar}
\usepackage{amsmath}
\usepackage{amssymb}
\usepackage{amsthm}
\usepackage{mathrsfs}
\usepackage{amsfonts}
\usepackage{bm}
\usepackage{algorithm}
\usepackage{algpseudocode}
\usepackage{url}
\usepackage{balance}
\usepackage{dblfloatfix}
\usepackage{flushend}
\usepackage{colortbl}
\usepackage{soul}
\usepackage{multicol} 
\usepackage{multirow}
\usepackage{pifont}
\usepackage{color}
\usepackage{alltt}
\usepackage[hidelinks]{hyperref}
\usepackage{enumerate}
\usepackage{siunitx}
\usepackage{breakurl}
\usepackage{enumitem}
\usepackage{epstopdf}
\usepackage{pbox}
\usepackage{multirow}

\usepackage{array,tabularx}
\usepackage{makecell}
\usepackage{subcaption}
\usepackage{cancel}
\usepackage{dsfont}
\usepackage{xcolor}
\usepackage{ctable}
\usepackage{kotex}
\usepackage{fancyhdr}
\usepackage{comment}
\usepackage{booktabs,caption}
\usepackage[flushleft]{threeparttable}
\usepackage[capitalize]{cleveref}
\usepackage[dvipsnames]{xcolor}

\usepackage{wrapfig}   

\newcommand{\rom}[1]{\uppercase\expandafter{\romannumeral #1\relax}}
\DeclareMathOperator*{\argmax}{arg\,max}

\newtheorem{theorem}{Theorem}[section]
\newtheorem{proposition}[theorem]{Proposition}
\newtheorem{lemma}[theorem]{Lemma}

\newtheorem{definition}{Definition}

\journal{Under review}

\begin{document}

	\begin{frontmatter}
	
		\title{Reinforcement Learning via Conservative Agent for Environments with Random Delays}
		
		\author[1]{Jongsoo Lee}
		\ead{jjongs@postech.ac.kr}
		          
		\author[1]{Jangwon Kim}
		\ead{jangwonkim@postech.ac.kr}
			
		\author[2]{Jiseok Jeong}
		\ead{wlwl9865@postech.ac.kr}
						
		\author[1]{Soohee Han\corref{cor1}}
		\ead{sooheehan@postech.ac.kr}

        \cortext[cor1]{Corresponding authors.}
		
		\affiliation[1]{organization={Department of Convergence IT Engineering, Pohang University of Science and Technology},
			addressline={77 Cheongam-ro},
			city={Nam-gu, Pohang-si},
			state={Gyeongbuk},
			postcode={36763},
			country={South Korea}}

        \affiliation[2]{organization={Department of Electrical Engineering, Pohang University of Science and Technology},
			addressline={77 Cheongam-ro},
			city={Nam-gu, Pohang-si},
			state={Gyeongbuk},
			postcode={36763},
			country={South Korea}}


		\begin{abstract}
          Real-world reinforcement learning applications are often subject to unavoidable delayed feedback from the environment. Under such conditions, the standard state representation may no longer induce Markovian dynamics unless additional information is incorporated at decision time, which introduces significant challenges for both learning and control. While numerous delay-compensation methods have been proposed for environments with constant delays, those with random delays remain largely unexplored due to their inherent variability and unpredictability. In this study, we propose a robust agent for decision-making under bounded random delays, termed the \textit{conservative agent}. This agent reformulates the random-delay environment into a constant-delay surrogate, which enables any constant-delay method to be directly extended to random-delay environments without modifying their algorithmic structure. Apart from a maximum delay, the conservative agent does not require prior knowledge of the underlying delay distribution and maintains performance invariant to changes in the delay distribution as long as the maximum delay remains unchanged. We present a theoretical analysis of conservative agent and evaluate its performance on diverse continuous control tasks from the MuJoCo benchmarks. Empirical results demonstrate that it significantly outperforms existing baselines in terms of both asymptotic performance and sample efficiency.
		\end{abstract}

		\begin{keyword}
            Markov decision process \sep random delay \sep reinforcement learning
		\end{keyword}
	
	\end{frontmatter}


\section{Introduction}

Reinforcement learning (RL) has achieved remarkable progress across a wide range of domains, from games \cite{dqn, RL-progress1, Game2, adaptive} to control systems \cite{Myung, intrinsic, traffic, building, initially, parkour1, parkour2, parkour3}. Mnih et al. \cite{dqn} demonstrated an RL agent capable of learning control policies directly from high-dimensional sensory inputs, achieving performance comparable to or exceeding that of human experts in Atari games. Wang et al. \cite{Game2} further presented an RL agent that defeats top professional players in the complex real-time strategy (RTS) game StarCraft II by combining imitation learning \cite{imitation} with RL. Within control systems, RL has been successfully applied to practical problems such as traffic signal control (TSC) for optimizing traffic flow and mitigating congestion in urban areas \cite{traffic}, and heating, ventilation, and air conditioning (HVAC) systems in commercial buildings for managing multi-zone precooling and improving energy efficiency \cite{building}. From a control-theoretic perspective, Wu et al. \cite{initially} proposed an RL framework for a discrete-time linear control system with multiple decision makers, modeled as a nonzero-sum multiplayer game, which approximates Nash-equilibrium solutions under an initial-excitation condition. Recent advances in RL have also enabled quadruped and humanoid robots to acquire highly agile and dynamic locomotion and to perform complex parkour-like maneuvers over challenging, unstructured terrain \cite{parkour1, parkour2, parkour3}.

Through a systematic trial-and-error process involving interactions with the environment, RL agents develop proficient decision-making capabilities that enable them to solve challenging problems, as evidenced by an expanding range of successful applications beyond the pivotal domains mentioned above \cite{RLHF, tokamak, finance, dense}. However, real-world RL applications often encounter challenges associated with delayed feedback, such as network latency in communication systems and response lags in robotic control systems \cite{NCS, robotic-delay, kaufmann}. Such delays can make the standard state representation insufficient to ensure the Markovian dynamics unless sufficient information is incorporated at decision time. If left unaddressed, this deficiency not only degrade the performance of RL agents but can also destabilize its behavior in dynamic systems \cite{hwangbo, mahmood}.

Existing research on addressing delay problems within the RL framework can be broadly categorized into two branches: the complete information method and the model-based method. The complete information method reconstructs the original state by augmenting it with delay-relevant historical information, thereby restoring Markovian dynamics in the presence of delays \cite{augmented}. Despite its solid theoretical foundation, its core strategy of augmenting the state can induce problems associated with sample complexity. To address this limitation, recurrent neural networks \cite{GRU} have been employed to embed the delay-relevant historical information into a compact hidden state, thereby recovering a Markovian state representation from partially observable dynamics without explicitly enlarging the state space \cite{Duel-RNN}. Alternatively, Kim et al. \cite{BPQL} introduces a representation for augmented values that are evaluated with respect to the original state space rather than the augmented one, which inherently mitigates the sample inefficiency. Wu et al. \cite{AD-SAC} alleviates performance degradation from this issue by leveraging auxiliary tasks with shorter delays to assist learning in tasks with longer delays. Wang et al. \cite{signal} takes a different approach that utilizes offline data to obtain time-calibrated information to avoid this sample complexity issue. More recently, Wu et al. \cite{VDPO} formulates delayed RL as a variational inference problem and proposes an iterative method that first learns a delay-free policy and then distills it into delayed environments via behavior cloning, achieving high sample efficiency.  

On the other hand, the model-based method aims to restore the Markovian dynamics by learning the underlying delay-free dynamics and then planning in delayed environments using the learned dynamics model \cite{mbs, At-Human, Delay-aware, acting}. For example, Derman et al. \cite{acting} approximates MDP dynamics model from transition samples collected in a delay-free environment and employs it in its delayed environment to infer non-delayed information through recursive one-step predictions. Similarly, Firoiu et al. \cite{At-Human} adopts recurrent neural networks to capture temporal dynamics. Despite facilitating more sample-efficient learning, their strong reliance on accurate dynamics modeling makes them vulnerable to model errors and stochasticity in environments; even slight prediction inaccuracies can accumulate, severely degrading policy performance and stability.

Although numerous delay-compensating methods have been proposed within the RL framework and have demonstrated their own promise, most of these approaches rely on the unrealistic assumption of constant delays or require explicit knowledge of the underlying delay distributions \cite{BPQL, AD-SAC, VDPO, Delay-aware, acting, DIDA}. Consequently, environments with random delays remain relatively underexplored compared to their constant-delay counterparts, mainly due to their inherent variability and unpredictability. This stochastic nature complicates the direct adaptation of conventional constant-delay methods, which often depend on structured and ordered historical information, and thus substantially limits their applicability in realistic settings involving random delays with unknown characteristics.

In this study, we propose an agent for robust decision-making under bounded random delays, termed the \textit{conservative agent}. This agent reformulates a random-delay environment into a constant-delay surrogate, providing a plug-and-play framework that allows any constant-delay method to be directly extended to random-delay environments without modifying its algorithmic structure. Apart from a maximum delay, the conservative agent does not require prior knowledge of the underlying delay distribution and demonstrates performance that is invariant to changes in the delay distribution as long as this maximum delay remains unchanged. This agent eliminates the need to estimate either individual delays or the delay distribution itself, effectively addressing the difficulties induced by the inherent unpredictability of random delays. We present a theoretical analysis of the conservative agent and evaluate its performance on diverse continuous control tasks from the MuJoCo benchmark \cite{mujoco}. Experimental results show that the proposed conservative RL algorithm consistently outperforms the state-of-the-art random-delay baselines in terms of both asymptotic performance and sample efficiency. The main contributions of this study are summarized as follows:

\begin{itemize}
    \item We propose a conservative agent for robust decision-making under bounded random delays. It reformulates a random-delay environment into a constant-delay surrogate, thereby providing a plug-and-play framework that allows conventional constant-delay methods to be directly applied to random-delay environments without algorithmic modification. Apart from a maximum delay, it does not require prior knowledge of the underlying delay distribution and exhibits performance that is invariant to changes in the delay distribution as long as this maximum delay remains unchanged.

    \item We derive the formal bounds on the performance gap between the normal and conservative agents under bounded random delays, which provide a guidance on when the conservative agent is well justified.

    \item We investigate the limitations of our approach and propose an effective mitigation strategy. Empirically, we confirm that the resulting framework achieves substantially better performance than state-of-the-art random-delay methods that operate directly in random-delay environments.
    
    \item We provide an empirical evidence supporting our theoretical findings and evaluate the proposed conservative RL algorithms on diverse continuous control tasks from the MuJoCo benchmark. Empirical results confirm that our algorithm significantly outperforms the state-of-the-art random-delay baselines in terms of both asymptotic performance and sample efficiency.      
    
\end{itemize}

\section{Preliminaries}

\subsection{Delay-free Reinforcement Learning} 

A Markov decision process (MDP) \cite{MDP} can be defined with a five-tuple $\mathcal{M} = (\mathcal{S}$, $\mathcal{A}$, $\mathcal{P}$, $\mathcal{R}$, $\gamma)$, where $\mathcal{S}$ and $\mathcal{A}$ represent the state and action spaces, $\mathcal{P}: \mathcal{S} \times \mathcal{A} \times \mathcal{S} \rightarrow [0, 1]$ is the transition kernel, $\mathcal{R}: \mathcal{S} \times \mathcal{A} \rightarrow \mathbb{R}$ is the reward function, and $\gamma \in (0, 1)$ is a discount factor. Additionally, a policy $\pi: \mathcal{S} \times \mathcal{A} \rightarrow [0,1]$ maps the state-to-action distribution. At each time step $t$, an agent observes a state $s_t \in \mathcal{S}$ from the environment, selects an action $a_t \in \mathcal{A}$ according to $\pi$, receives a bounded reward $r_t = \mathcal{R}(s_t, a_t) \in [0, R_\text{max}]$, and then observes the next state $s_{t+1}  \in \mathcal{S}$. The agent repeats this process to find an optimal policy $\pi^{*}$ that maximizes the expected return over a finite or infinite-horizon $H$, which is given as: 
\begin{align}
\pi^* = \argmax_{\pi} \; \mathbb{E}\left[\sum_{k=0}^{H-1}\gamma^{k}r_{k+1} \; \Big\vert \; \pi, \rho \right],  
\end{align}where $\rho$ denotes the initial state distribution under $\pi$. The value functions are then defined as:
\begin{align}
V^{\pi}(s) =  \mathbb{E}\left[\sum_{k=0}^{H-1}\gamma^{k}r_{t+k}  \Big\vert  s_{t}=s, \pi \right], \;\; Q^{\pi}(s, a) =  \mathbb{E}\left[\sum_{k=0}^{H-1}\gamma^{k}r_{t+k}  \Big \vert s_{t}=s, a_{t}=a, \pi \right],
\end{align} where $V^{\pi}(s)$ is the state-value function denoting the expected return starting from state $s$ under the policy $\pi$, and $Q^{\pi}(s, a)$ is the $Q$-value function representing the expected return starting from state $s$, taking action $a$, and thereafter following the policy $\pi$. Notably, the dynamics governing MDPs assume the Markov property, implying that the complete probability distribution in the dynamics is fully determined by the present state and action, independent of their histories. However, delayed environmental feedback can make the standard state representation insufficient to ensure Markovian dynamics, which can significantly degrade the performance of RL agents and even cause complete failure in both learning and control \cite{POMDP2}.


\subsection{Delayed Reinforcement Learning}

In environments with delays, modeled as delayed MDPs, the state may not be observed by the agent immediately (observation delay, $\Delta T_s$), a lag may exist in determining an action (inference delay, $\Delta T_i$), and the effect of the action may also be delayed (execution delay, $\Delta T_e$). Such delays force the agent to make decisions based on outdated information and prevent timely actions, disrupting the overall learning process. Fortunately, only the total delay $\Delta T = \Delta T_s + \Delta T_i + \Delta T_e$ matters for decision-making \cite{signal}, allowing different types of delays to be treated equivalently and simplifying the analysis. Based on this rationale, we concentrate on observation delays without loss of generality. 

In particular, random delays can cause states to be observed out of their chronological order. To enable formal reasoning in such environments, we distinguish among state-generation, state-observation, and state-usage events and assume that an observed state can be used for decision-making only after all previously generated states have been used. This yields a tractable framework for analyzing the decision-making process under bounded random delays. In addition, following \cite{augmented, BPQL, DelaysInRL}, we assume that reward feedback is received concurrently with state feedback in order to avoid collecting partial information from the reward signal.

\begin{definition} \label{def-1}
    A state is considered generated when the environment responds to an agent's action, observed when the resulting state information reaches the agent, and used when the agent utilizes it to make a decision by feeding it into the policy.
\end{definition}

\subsection{Complete Information Method} \label{Section-3}
The complete information method is often preferred for compensating delayed effects in delayed MDPs, as it restores Markovian dynamics in the presence of delays and enables policy learning with conventional RL algorithms. As shown by \cite{augmented}, any delayed MDP can be transformed into an equivalent delay-free MDP via state augmentation, while preserving optimal policies. In this section, we first formalize constant-delay MDPs and then extend this framework to random-delay MDPs.

\subsubsection{Constant-delay MDP} \label{section-3.1}

A constant-delay MDP can be defined as an extended tuple $\mathcal{M}^+ = (\mathcal{M}, \Delta)$, where $\Delta \in \mathbb{N}$ denotes a fixed observation delay. This can be reduced to an equivalent delay-free MDP $\mathcal{M}_\Delta = (\mathcal{X}_\Delta$, $\mathcal{A}$, $\mathcal{P}_\Delta$, $\mathcal{R}_\Delta, \gamma)$, where $\mathcal{X}_\Delta = \mathcal{S} \times \mathcal{A}^{\Delta}$ is the augmented state space with $\mathcal{A}^{\Delta}$ being the Cartesian product of $\mathcal{A}$ with itself for $\Delta$ times, $\mathcal{P}_\Delta : \mathcal{X}_\Delta \times \mathcal{A} \times \mathcal{X}_\Delta \rightarrow [0, 1]$ and $\mathcal{R}_\Delta: \mathcal{X}_\Delta \times \mathcal{A} \rightarrow \mathbb{R}$ denote the augmented transition kernel and the augmented reward function, respectively. In addition, the augmented state-based policy $\pi_\Delta : \mathcal{X}_\Delta \times \mathcal{A} \rightarrow [0, 1]$ maps the augmented state-to-action distribution. Specifically, the augmented state $ x \in \mathcal{X}_\Delta$ at time $t$ is defined as: 
\begin{align} \label{eq:augmented-state-in-CDMDP}
x_{t} &= (s_{t-\Delta}, a_{t-\Delta}, a_{t-\Delta+1}, \dots, a_{t-1}), \quad \forall t > \Delta,    
\end{align}where $s_{t-\Delta}$ is the most recently observed state and $(a_i)^{t-1}_{i=t-\Delta}$ is the sequence of historical actions. Then the augmented transition kernel is defined as:
\begin{align}
    \mathcal{P}_\Delta(x_{t+1} \mid x_t, a_t) = \mathcal{P}(s_{t-\Delta+1} \mid s_{t-\Delta}, a_{t-\Delta})\delta_{a_t}(a'_t)\prod^{\Delta-1}_{i=1}\delta_{a_{t-i}}(a'_{t-i}),
\end{align} where $\delta$ denotes the Direct-delta distribution. Given the augmented state $x_{t}$, the current state $s_t$ can be inferred based on a belief $b_\Delta(s_t \mid x_t)$ \cite{DelaysInRL}, representing the probability of $s_t$ given $x_t$
\begin{align}
b_\Delta(s_t \mid x_t) = \int_{\mathcal{S}^{\Delta-1}}\mathcal{P}(s_t \mid s_{t-1}, a_{t-1}) \prod^{t-2}_{i=t-\Delta}\mathcal{P}(s_{i+1} \mid s_{i}, a_{i})ds_{i+1}.
\end{align} Learning this belief enables the agent to infer non-delayed information. Note that since $s_t$ is not explicitly observed at current time $t$, the augmented reward is defined as the expectation under this belief:
\begin{align}
    \mathcal{R}_\Delta(x_t, a_t) = \mathbb{E}_{b_\Delta}\left[\mathcal{R}(s_t, a_t)\right].
\end{align}

\subsubsection{Random-delay MDPs} \label{section-3.2}

Extending the constant-delay MDP framework, a random-delay MDP can be defined as $\mathcal{M}_{\lambda} = (\mathcal{M}, \lambda, \tau)$, where the delay is no longer a fixed value. Instead, it is sampled from a delay distribution $\lambda$ supported on a finite integer set $\Lambda = \{1, 2, \dots, \Delta_\text{max}\}$ at each time step $t$. The time-indexing function $\tau: \mathcal{S} \rightarrow \mathbb{N}$ maps each observed state to the time at which it becomes available to the agent for decision-making. Under this definition, the augmented states can be defined as follows:

\begin{definition}
Let $s_{n}$ and $\Delta^{(n)}$ denote the state generated at time step $n$ and the sampled delay, respectively, and let $(a_i)^{t-1}_{i=n}$ denote the sequence of historical actions. Suppose the state $s_{n}$ is the most recent usable state at time step $t > n$, then the augmented state is defined as: 
\begin{align} \label{eq:RDMDP}
\tilde{x}_t = (s_n, a_n, a_{n+1}, \dots, a_{t-1}), \quad \text{for some} \; t \in \left[\tau(s_n), n+\Delta_\text{max}\right] \cap \mathbb{Z},
\end{align} where $n+\Delta_\text{max}$ denotes the maximally delayed time step. Assuming that $\tau(s_{n}) = t$, the subsequent augmented state is defined as: 
\begin{align} \label{eq:augmented-state}
\tilde{x}_{t+1} = \begin{cases}
    (s_{n+1}, a_{n+1}, a_{n+2}, \dots, a_t), &\text{if } n + \Delta^{(n+1)} \leq \tau(s_n), \\
    (s_n, a_n, a_{n+1}, \dots, a_t), &\text{otherwise}.
    \end{cases}
\end{align}
where $\mathbb{P}\left(n +\Delta^{(n+1)}\leq\tau(s_n)\right)$ denotes the probability that the subsequent state $s_{n+1}\sim\mathcal{P}(\cdot \mid s_{n}, {a_n})$ becomes available by this time. Notably, the state continues to be used for decision-making until the subsequent state becomes available to use. 
\end{definition} 

Note that the dimension of the augmented state either remains constant or increases by one at each time step. This implies that it can grow without bound in infinite-horizon MDPs if delays are not assumed to be bounded. To address this issue, \cite{augmented} introduced a method called \textit{freeze}, in which the agent takes no action until the timely state $s$ is observed whenever the augmented state reaches its maximum allowable dimension. Once such a state $s$ is observed, the augmented state is reset to $\tilde{x} = (s)$, and the decision process resumes. However, the performance of the freeze agent is known to be highly task-dependent and unstable, since the agent disregards environmental changes during its inactive periods.

\section{Conservative Reinforcement Learning} \label{Section-5}

In this section, we propose a conservative agent and demonstrate how it reformulates the random-delay environment into its constant-delay surrogate. We then present theoretical results that provide guidance on when the conservative agent is well justified. Finally, we investigate a limitation of the conservative approach and subsequently introduce an effective mitigation strategy.

\subsection{Conservative Agent} \label{lazy-agent-section}

We introduce an agent that follows a \textit{conservative} decision-making strategy: it makes decisions on each observed state only at the maximally delayed time step. Concretely, for any state $s_n$ with delay $\Delta^{(n)}$, this conservative agent assumes the state becomes available for decision-making at time step $n + \Delta_\text{max}$, that is, 
\begin{align}
 \tau(s_{n}) = n+\Delta_{\text{max}}, \quad \forall n > 0.
\end{align} Therefore, each state is used for decision-making exactly $\Delta_{\text{max}}$ time steps after its generation. This strategy obviates the need to estimate individual delays or the delay distribution itself, yielding a delay-agnostic agent. Moreover, as long as the maximum delay remains unchanged, it guarantees that performance is invariant to changes in the delay distribution, since it always constructs the same constant-delay surrogate. Formally, the conservative agent takes the same shift structure at every time step such that
\begin{align} 
\tilde{x}_{t} &= (s_{t-\Delta_\text{max}}, a_{t-\Delta_\text{max}}, a_{t-\Delta_\text{max}+1}, ..., a_{t-1}), \quad \forall t>\Delta_\text{max}. 
\end{align} This formulation matches the definition of augmented states in constant-delay MDPs with delay $\Delta = \Delta_\text{max}$, which can be further reduced to an equivalent delay-free MDP. This implies that conventional constant-delay methods can be extended to random-delay MDPs without modifying the underlying algorithmic structure or explicitly handling delay-induced uncertainties. From this, we say that the conservative agent constructs a \textit{constant-delay surrogate} of a random-delay MDP, as stated below.


\begin{proposition} \label{proposition}
Let $\lambda$ be a delay distribution supported on the set $\Lambda = \{1, 2, \dots, \Delta_\text{max}\}$. If the agent follows the conservative decision-making strategy that assumes $\tau(s_{n}) = n+\Delta_{\text{max}}$ for all $n > 0$, then a random-delay MDP can be reformulated as a constant-delay surrogate with fixed delay $\Delta = \Delta_\text{max}$, which can be further reduced to an equivalent delay-free MDP.
\end{proposition}

\begin{proof}
    See \ref{proof-proposition-1}. 
\end{proof}

\begin{figure*}[!h]
    \centering
    \includegraphics[width=1\textwidth]{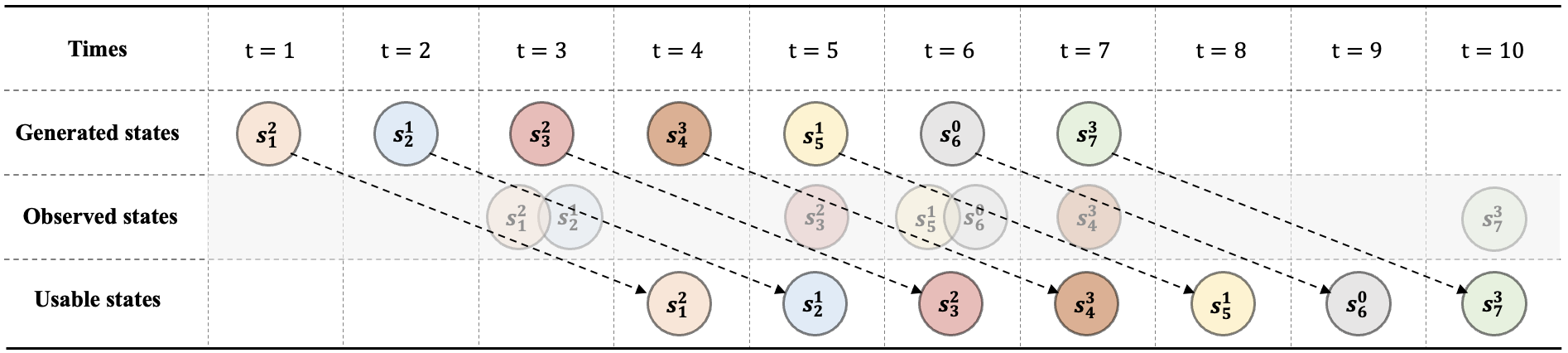}
\caption{A visual example illustrating the conservative decision-making process under random delays with $\Delta_\text{max}=3$, where the subscripts denote state-generation times and the superscripts indicate the corresponding delays. Despite simultaneous or out-of-order observations, each state is used for decision-making exactly $\Delta_\text{max}$ time steps after its generation.} \label{lazy-framework} 
\end{figure*} 

To support our claim, we present empirical results in Section \ref{Experiment}, showing that the performance of the conservative agent in random-delay MDPs is invariant to the distributional changes and is \textit{nearly identical} to that of the normal agent in constant-delay MDPs with fixed delay $\Delta = \Delta_\text{max}$. A visual representation is provided in Fig.~\ref{lazy-framework}. Note that both the conservative and freeze approaches \cite{augmented} share a common foundation in that they are built upon the classical complete information method and are designed to handle random delays. Nevertheless, these two approaches rely on different assumptions and employ conceptually different control mechanisms. The freeze agent can be viewed as a state-representation-level control mechanism that prevents the dimension of the augmented state from growing without bound under delays modeled as discrete, finite-valued random variables. In contrast, the conservative agent can be regarded as a policy-level control mechanism that regularizes the timing of state usage under bounded random delays, while still ensuring that the augmented state representation remains fixed-dimensional. The details are provided in \ref{freeze-agent}.

\subsection{Theoretical Analysis} \label{theory}

We derive a theoretical bound on the performance difference between the normal agent operating directly in a random-delay MDP and the conservative agent operating in the corresponding constant-delay surrogate. To facilitate theoretical analysis, we introduce a meta-policy $\pi_{\text{meta}}$ that selects delay-specific constant-delay policies under bounded random delays. In particular, let $\{\pi^*_\Delta\}_{\Delta\in\Lambda}$ be the set of optimal constant-delay policies for constant-delay MDPs with fixed delay $\Delta \in \Lambda$ and let their performance be defined as:
\begin{align}
    J^*_{\Delta} \;  \triangleq \;  \mathbb{E}_{\rho_{\Delta}}[V^{\pi^*_\Delta}(x)],  
\end{align} where $\rho_\Delta$ denotes the initial augmented state distribution under $\pi^*_{\Delta}$. Note that the performance of conservative agent matches that of $\pi^*_\Delta$ with $\Delta = \Delta_\text{max}$ denoted by $J^*_\text{max}$. At execution time, after observing the state and its realized delay $\Delta$, the meta-agent forms the augmented state $x \in \mathcal{X}_{\Delta}$ and chooses the optimal constant-delay policy $\pi^*_\Delta$ to determine its action. Formally, for any $a \in \mathcal{A}$, the meta-policy is defined as
\begin{align}
 \pi_{\text{meta}}(a \mid x,\Delta) \; \triangleq \; \pi^*_{\Delta}(a \mid x).
\end{align} For analytical tractability, we assume that the performance of the meta-policy under random delays can be characterized as the weighted average of the constituent performances such that
\begin{align}
    J_{\text{meta}} \; \triangleq \; \mathbb{E}_{\lambda}[J^*_{\Delta}],
\end{align}where $\lambda$ denotes the delay distribution supported on $\Lambda$. We begin by comparing the performance of optimal constant-delay policies, as stated in the Lemma below.

\begin{lemma} [Theorem 4.3.1 in \cite{DelaysInRL}] \label{theorem-1}
     Let there be two constant-delay MDPs with fixed delays $\Delta, \Delta' \in \Lambda$, respectively, and let $J^*_{1}$ and $J^*_{2}$ denote their respective performance under optimal constant-delay policies. Then if $\Delta < \Delta'$, it satisfies $J^*_{1} \geq J^*_{2}$.
\end{lemma} This implies that the optimal constant-delay performance is monotonically non-increasing in the delay. Building on this insight, we derive a formal bound on the performance gap between the meta-policy and the conservative policy, as stated below.

\begin{theorem}\label{theorem-2}
Let $\gamma\in(0,1)$ and $r_t\in[0,R_{\max}]$ for all $t\ge 0$. Then for any $\Delta, \Delta' \in \Lambda$,
\begin{align}
    \left|J^*_{\Delta} - J^*_{\Delta'}\right|  \leq  \frac{\gamma^{\min(\Delta, \Delta')}}{1 - \gamma}R_\text{max}\left|\Delta-\Delta'\right|.
\end{align} Moreover, it extends to 
\begin{align}
    0 \leq J_\text{meta} - J^*_\text{max} \leq \frac{\gamma^{\Delta_\text{min}}}{1 - \gamma}R_\text{max}\left(\Delta_\text{max} - \mathbb{E}_{\lambda}\left[\Delta\right]\right), \label{bound-1}
\end{align} where $\Delta_\text{min}$ and $\Delta_\text{max}$ denote the minimum and maximum supports of the delay distribution $\lambda$.
\end{theorem}

\begin{proof}
    See \ref{proof-theorem-1}.
\end{proof}

In Theorem~\ref{theorem-2}, we first establish a bound on the performance gap between optimal constant-delay policies with different fixed delays. Building on this result, we derive a distribution-dependent bound on the performance difference between the meta-policy and the conservative policy under bounded random delays. This bound suggests that the conservative approach is well justified when the deviation between the expected delay and the maximum delay is reasonably small, or when the minimum delay is relatively large so that all delays are concentrated near the maximum delay. Moreover, we can further obtain a distribution-independent performance bound, as stated below.

\begin{theorem}\label{theorem-3}
Let $\gamma\in(0,1)$ and $r_t\in[0,R_{\max}]$ for all $t\ge 0$. Then,
\begin{align}
0 \leq J_\text{meta} - J^*_{\text{max}}\leq\frac{R_\text{max}}{1 - \gamma}(1 - \gamma^{\Delta_\text{max}}),
\end{align}where $\Delta_\text{max}$ denotes the maximum support of the delay distribution $\lambda$.
\end{theorem}

\begin{proof}
    See \ref{proof-theorem-2}.
\end{proof}


Intuitively, this result implies that the performance of meta-policy and conservative policy achieves nearly identical performance when $\Delta_{\text{max}}$ is small. Although this bound is quite pessimistic as it deliberately ignores structural properties of the underlying MDP, it remains valid even when the delay distribution is unknown or varies over time. Thus, it provides a robust guideline for when the conservative approach is well justified. Notably, the conservative agent is delay-agnostic, whose performance is inherently independent of changes in the delay distribution and can be applied without any prior knowledge of this distribution. By contrast, the meta-policy is explicitly distribution-dependent and requires the knowledge of the delay distribution in order to be constructed. These observations strongly support the effectiveness of our conservative approach. In the following section, we investigate a limitation of the conservative approach and subsequently introduce an effective mitigation strategy.

\subsection{Conservative-BPQL} \label{conservative-BPQL}

Previously, we showed that the conservative agent reformulates a random-delay MDP into its constant-delay surrogate, which can be further reduced to an equivalent delay-free MDP. Moreover, we demonstrated that this strategy is well justified when the deviation between the expected and maximum delays is small, or when the maximum delay itself is small. However, the conservative agent raises a concern about sample complexity, since it takes the worst-case policy. Concretely, the sample complexity of $Q$-learning for the augmented state space $\mathcal{X}_\Delta = \mathcal{S} \times \mathcal{A}^{\Delta}$ is given by
\begin{align}
 O\Big(\frac{\log (\vert \mathcal{X}_\Delta\vert\vert\mathcal{A} \vert)}{\epsilon^{2.5}(1 - \gamma)^5}\Big) = O\Big(\frac{\log \vert \mathcal{S}\vert + (\Delta + 1) \log \vert\mathcal{A} \vert}{\epsilon^{2.5}(1 - \gamma)^5}\Big),
\end{align}
which characterizes the number of samples required for $Q$-learning to converge to an $\epsilon$-optimal with high confidence \cite{Speed-QL, AD-SAC}. This implies that the sample complexity of the conservative agent becomes larger as the maximum delay grows, resulting in severe sample inefficiency \cite{acting, BPQL}. Therefore, although the conservative agent eliminates the core difficulties caused by unpredictable delays, mitigating its sample complexity overhead is crucial to prevent unintended performance degradation. Fortunately, several effective constant-delay methods have already been proposed to tackle this issue, with a belief projection-based $Q$-learning (BPQL) \cite{BPQL} being a representative example. Motivated by this, we integrate the conservative agent into the BPQL framework, yielding the conservative-BPQL algorithm summarized in \ref{Appendix-algorithm}. 

In detail, BPQL is an actor-critic algorithm built upon the soft actor-critic (SAC) algorithm \cite{sac}, which learns the critic using an approximate value function over the augmented state given by 
\begin{align}  
\tilde{Q}^{\pi_{\Delta}}(\tilde{x}_t, a_t) &= \mathbb{E}_{b_{\Delta}}\left[ Q^{\pi_{\Delta}}_{\beta}(s_t, a_t)\right] + \zeta^{\pi_\Delta}(\tilde{x}_t, a_t), \label{action-projection}
\end{align} where $\Delta = \Delta_\text{max}$, $\tilde{Q}^{\pi_\Delta}$ and $Q^{\pi_\Delta}_{\beta}$ denote the augmented state-based $Q$-value function and beta $Q$-value function, respectively, and $\zeta^{\pi_{\Delta}}$ represents the projection residual. Notably, $Q^{\pi_{\Delta}}_{\beta}$ is evaluated with respect to the original state space rather than the augmented one, thereby inherently mitigating the learning inefficiencies. Practically, the beta $Q$-value function $Q^{\pi_\Delta}_{\beta}$ and the augmented state-based policy $\pi_\Delta$ are parameterized by $\theta$ (beta-critic) and $\phi$ (actor), respectively. These parameters are updated iteratively to minimize the following objective functions; for the beta-critic: 
\begin{align}
\mathcal{J}_{Q_{\beta}}(\theta) = \mathbb{E}_{\mathcal{D}} \bigg[\frac{1}{2}\Big(Q^{\pi_{\phi}}_{\theta, \beta}(s_t, a_t) - \mathcal{R}(s_t, a_t) - \gamma\mathbb{E}_{a_{t+1} \sim \pi_{\phi}} [Q_{\theta',{\beta}}^{\pi_{\phi}}(s_{t+1}, a_{t+1}) - \alpha\log\pi_{\phi}(a_{t+1} \mid \tilde{x}_{t+1})]\Big)^{2}\bigg]
\end{align} and for the actor:
\begin{align}
\mathcal{J}_{\pi}(\phi) = \mathbb{E}_{\mathcal{D}} \Big[\mathbb{E}_{a_t \sim {\pi}_{\phi}} [ \alpha \log \pi_{\phi}(a_t \mid \tilde{x}_t)- Q^{\pi_\phi}_{\theta, \beta}(s_t, a_t) ]\Big],
\end{align}where $\mathcal{D}$ is a replay buffer, $\alpha$ is a temperature, and $\theta'$ are the parameters of the target beta-critic \cite{dqn, sac, td3}. 

Despite demonstrating notable performance under fixed delays, the BPQL framework cannot be directly applied to random-delay MDPs due to its inherent design limitations. As discussed above, the conservative agent provides a plug-in mechanism that enables the BPQL framework to be extended to random-delay settings. This justifies the use of BPQL in random-delay MDPs. It is worth emphasizing that, although we specifically instantiate the conservative agent within the BPQL framework, any constant-delay method can, in principle, be extended to random-delay MDPs by leveraging our agent. To demonstrate this generality, we develop a second conservative RL algorithm built upon another state-of-the-art constant-delay method, variational delayed policy optimization (VDPO) \cite{VDPO}. The corresponding results are presented in Section~\ref{Ablation}.

\section{Experiments} \label{Experiment}

In this section, we provide empirical validation of the theoretical analysis presented in Section~\ref{theory} and evaluate the overall performance of the proposed conservative RL algorithm in comparison with state-of-the-art random-delay algorithms. In addition, we conduct an ablation study to examine the general applicability of the conservative agent and to quantify its computational overhead.

\subsection{Benchmarks and Baseline Algorithms}

We evaluated the conservative agent on diverse continuous control tasks from the MuJoCo benchmark under random delays with $\Delta_{\text{max}} \in \{5, 10, 20\}$, together with the state-of-the-art random-delay algorithms. To validate our theoretical analysis, we first model the random delay using a Poisson distribution with various rate parameters to assess robustness to changes in both the distributional form and the expected delay. We then evaluate the overall performance of the baseline algorithms under uniformly distributed random delays. Details of the benchmark environments and the experimental setup are provided in \ref{Experimental-detail}.

The following algorithms were included as baselines in our experiments: normal SAC \cite{sac}, delayed-SAC \cite{acting}, DC/AC \cite{DCAC}, and state augmentation-MLP \cite{signal}. The normal SAC serves as a naive baseline that selects actions for currently usable states in a memoryless basis and performs `no-ops' otherwise, without concerning the disruption of the Markov assumption. Delayed-SAC is a variant of delayed-$Q$ \cite{acting}, adapted by Kim et al. \cite{BPQL} for continuous spaces. It constructs an approximate dynamics model to infer unobserved states through recursive one-step predictions. DC/AC is another variant of SAC that incorporates off-policy multi-step value estimation with a partial trajectory resampling strategy, significantly improving sample efficiency and demonstrating remarkable performance under random delays. Lastly, state augmentation-MLP is the most recent algorithm that leverages time-calibrated information obtained from offline datasets. Combined with auxiliary techniques, it has shown strong empirical performance under random delays.

\subsection{Validation for Theoretical Analysis} \label{validation}

\begin{figure*}[!h] 
    \centering
    \includegraphics[width=1.0\textwidth]{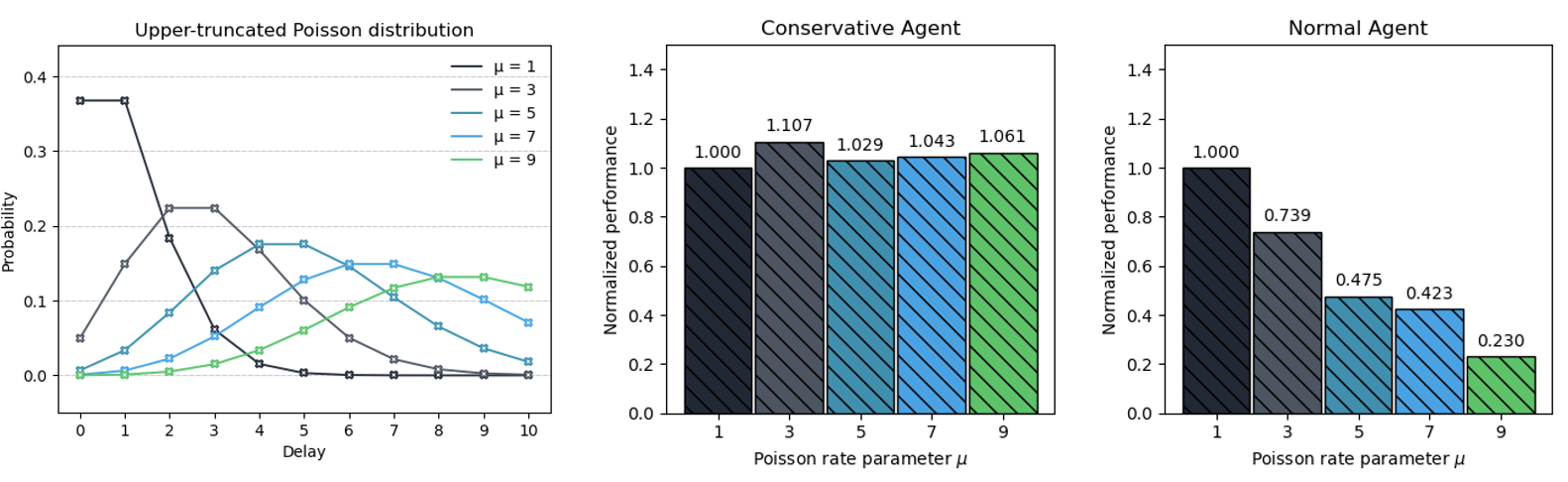}
    \caption{Normalized performance of the conservative agent and the normal agent in HalfCheetah-v3 task in MuJoCo benchmark under an upper-truncated Poisson delay distribution with rate parameter $\mu \in \{1, 3, 5, 7, 9\}$ shown on the left. The performance of each agent is averaged over 10 random seeds and normalized to its respective $\mu = 1$ baseline. While the normalized performance of the conservative agent is invariant to $\mu$, that of the normal agent deteriorates as $\mu$ increases.} \label{validation-fig}
\end{figure*} 

We first verify that the performance of the conservative agent is invariant to the changes in the underlying delay distribution and that the deviation between the expected delay and the maximum delay is closely related to the extent of its performance degradation when operating within the corresponding constant-delay surrogate. The delay distribution $\lambda$ is modeled as an upper-truncated Poisson distribution with probability mass function
\begin{align}
    \mathbb{P}(\Delta = k) = \frac{\mathrm{Pois}(k, \mu)}{\sum^{\Delta_\text{max}}_{j = 0}\mathrm{Pois}(j, \mu)}, \quad k = 0, 1, \dots, \Delta_\text{max},
\end{align} where $\mathrm{Pois}(k, \mu) \triangleq e^{-\mu}\mu^{k}/k!$ with $\mu>0$ being the Poisson rate parameter, equal to the mean. Note that the mean of untruncated Poisson distribution $\mu$ generally differs from the mean of the upper-truncated Poisson distribution $\tilde{\mu} = \mathbb{E}_{\lambda}[\Delta]$, but it satisfies $\tilde{\mu} \rightarrow \mu$ as $\Delta_\text{max} \rightarrow \infty$. We vary the Poisson rate parameter $\mu$ and report the resulting performance changes in Fig. \ref{validation-fig}. 

The results confirm that the performance of conservative agent remains invariant to $\mu$ for a fixed maximum delay, indicating that our agent is inherently robust to the distributional changes. On the contrary, the normal agent that operates directly in random-delay MDPs exhibited highly distribution–dependent performance. Formally, as noted in Theorem~\ref{theorem-2}, we have
\begin{align}
    J_\text{meta}  &\leq J^*_\text{max} + L(\Delta_\text{max} - \tilde{\mu}) = \underbrace{J^*_\text{max} + L\cdot\Delta_\text{max}}_{\text{constant}} - L \cdot \tilde{\mu},
\end{align} where $L = \gamma^{\Delta_\text{min}}R_\text{max}/(1-\gamma)$ and  $\tilde{\mu} = \mathbb{E}_{\lambda}[\Delta]$. This implies that the normal agent admits a linearly decreasing upper bound in $\tilde{\mu}$, and that $J_\text{meta} \rightarrow J^*_\text{max}$ as $\tilde{\mu} \rightarrow \Delta_\text{max}$, since $0 \leq J_\text{meta} - J^*_\text{max}$. Thus, operating within a constant-delay surrogate may degrade performance, yet the use of conservative agent can be justified when the deviation between the expected delay and the maximum delay is reasonably small. Moreover, when $\Delta_\text{max}$ is small itself, the extend of performance degradation is limited since $\tilde{\mu} \leq \Delta_\text{max}$. These observations closely align with the theoretical results presented in Section~\ref{theory}.

\subsection{Mitigating the Sample Complexity Issue}

\begin{figure*}[!h] 
    \centering
    \includegraphics[width=1.0\textwidth]{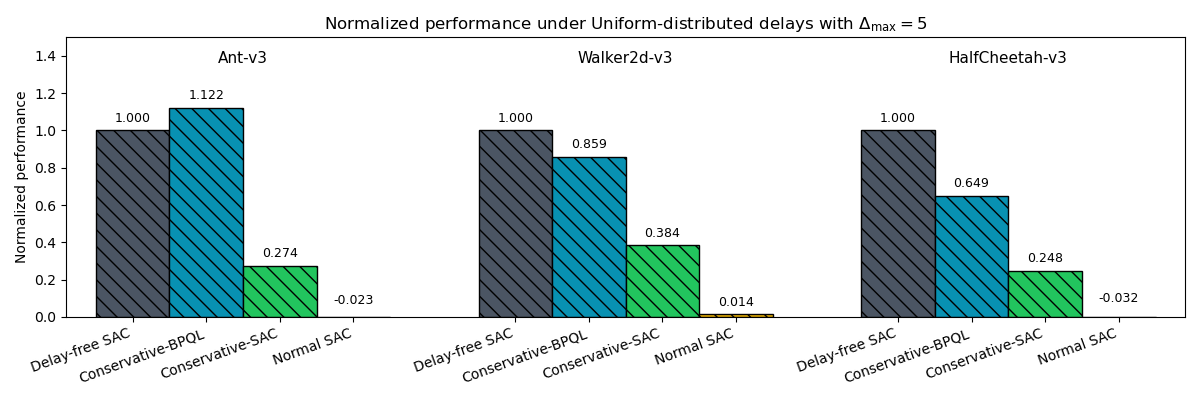}
    \caption{Normalized performance of the conservative agents, with (Conservative-BPQL) and without (Conservative-SAC) mitigation of the sample complexity issue, in MuJoCo environments under a uniform delay distribution with $\Delta_\text{max} = 5$. The results are averaged over five random seeds and normalized to the delay-free baseline (delay-free SAC).} \label{validation-fig-2}
\end{figure*} 

As previously pointed out, the conservative agent may introduce a sample complexity issue. To examine the importance and effectiveness of alleviating this side-effect, we evaluate the conservative agent without employing the BPQL technique, which we refer to as conservative-SAC. As presented in Fig.~\ref{validation-fig-2}, conservative-BPQL outperforms conservative-SAC across all evaluated tasks and even surpasses the delay-free baseline in the Ant-v3 environment. These results underscore the importance of mitigating the sample complexity issue for the conservative agent. Based on these empirical findings, we anticipate that our agent can achieve highly competitive performance under random delays when combined with a constant-delay method that effectively addresses the sample complexity issue. In the subsequent section, we compare the overall performance of our conservative RL algorithm with that of state-of-the-art random-delay algorithms.

\subsection{Performance Comparison} \label{comparison}

\begin{table*}[!h] 
\caption{Results on the MuJoCo benchmarks under random delays with $\Delta_{\text{max}} \in \{5, 10, 20\}$. Each algorithm was evaluated for one million time steps with five random seeds, where the standard deviations of average returns are denoted by $\pm$. The best performance under random delays is colored in {\color{blue}blue} and the performance under fixed delay $\Delta = \Delta_\text{max}$ is colored in {\color{red}red}.} \label{result-table}
\centering
\renewcommand{\arraystretch}{1.5}
\resizebox{\textwidth}{!}{
\begin{tabular}{cccccccc}
\Xhline{2\arrayrulewidth}
\multicolumn{2}{c}{\textbf{Environment}} & \multirow{2}{*}{Ant-v3} &  \multirow{2}{*}{HalfCheetah-v3} & \multirow{2}{*}{Hopper-v3} & \multirow{2}{*}{Walker2d-v3} & \multirow{2}{*}{Humanoid-v3} & \multirow{2}{*}{InvertedPendulum-v2} \\ 

\cline{1-2} 
$\Delta_\text{max}$ & \textbf{Algorithm}  & \\

\Xhline{1.5\arrayrulewidth}
\multirow{2}{*}{$\cdot$}  & Random policy & $-58.7_{\pm4}$ & $-285.01_{\pm3}$ & $18.6_{\pm2}$ & $1.9_{\pm1}$ & $121.9_{\pm2}$ & $5.6_{\pm1}$ \\
& Delay-free SAC             & $3279.2_{\pm180}$ & $8608.4_{\pm57}$ & $2435.2_{\pm23}$ & $3305.5_{\pm234}$ & $3228.1_{\pm410}$ & $964.3_{\pm29}$ \\
\hline
\multirow{6}{*}{5}  & Normal SAC             & $-76.6_{\pm4}$ & $-279.5_{\pm5}$ & $89.2_{\pm10}$ &                                                            $44.7_{\pm21}$ & $403.9_{\pm5}$ & $32.2_{\pm2}$ \\

                    & DC/AC                  & $907.5_{\pm90}$ & $2561.8_{\pm92}$  & $1931.6_{\pm192}$ &                                  $2079.3_{\pm122}$ & $2798.4_{\pm452}$ & $854.7_{\pm30}$ \\

                    & Delayed-SAC             & $986.4_{\pm128}$ & $4569.4_{\pm88}$ & {\color{blue}{$2200.4_{\pm190}$}} & $1910.1_{\pm247}$ & $418.9_{\pm126}$ & {\color{blue}{$964.2_{\pm15}$}} \\
                    
                    & State Augmentation-MLP    &  $916.1_{\pm621}$ &  $2442.1_{\pm79}$ &  $1946.1_{\pm144}$ &  $2144.8_{\pm361}$ & $493.3_{\pm18}$ &  $283.1_{\pm27}$ \\
                    
                    & Conservative-BPQL ({\color{blue}ours}) & {\color{blue}{$3679.8_{\pm167}$}} & {\color{blue}{$5583.9_{\pm169}$}} & $2174.1_{\pm155}$ & {\color{blue}{$2843.2_{\pm272}$}} & {\color{blue}{$3157.7_{\pm292}$}} & ${958.8}_{\pm14}$ \\
                    \cline{2-8}
                    & BPQL ({\color{red}constant}) & {\color{red}$3761.9_{\pm112}$} & {\color{red}{$5212.7_{\pm41}$}} & {\color{red}{$2136.3_{\pm158}$}} & {\color{red}{$2577.4_{\pm157}$}} & {\color{red}{$3194.9_{\pm374}$}} & {\color{red}{$955.9_{\pm28}$}} \\
\hline
\multirow{6}{*}{10} & Normal SAC                              & $-84.6_{\pm9}$ & $-278.6_{\pm6}$ & $28.1_{\pm6}$ & $40.9_{\pm4}$ & $354.5_{\pm12}$ & $31.3_{\pm1}$ \\
                
                    & DC/AC             & $342.9_{\pm34}$ & $1824.5_{\pm111}$ & $1262.4_{\pm261}$ & $1492.5_{\pm133}$ & $1023.8_{\pm359}$ & $4.9_{\pm0}$ \\
               
                    & Delayed-SAC             & $966.9_{\pm180}$ & $2563.8_{\pm215}$ & $1878.5_{\pm176}$ & $1264.6_{\pm233}$ & $289.6_{\pm108}$ & {\color{blue}{$947.6_{\pm36}$}} \\
                    
                    & State Augmentation-MLP   &  $779.4_{\pm554}$ &  $1926.3_{\pm197}$ &  $1286.7_{\pm124}$ &  $1148.3_{\pm283}$ & $470.3_{\pm11}$ &  $58.7_{\pm4}$ \\
                    
                    & Conservative-BPQL ({\color{blue}ours})  & {\color{blue}{$2744.5_{\pm112}$}} & {\color{blue}{$4511.8_{\pm143}$}} & {\color{blue}{$2300.9_{\pm164}$}} & {\color{blue}{$2122.3_{\pm292}$}} & {\color{blue}{$2820.5_{\pm348}$}} & $936.9_{\pm38}$ \\      
                    \cline{2-8}
                    & BPQL ({\color{red}constant}) & {\color{red}{$2831.9_{\pm103}$}} & {\color{red}{$4282.2_{\pm203}$}} & {\color{red}{$2129.2_{\pm184}$}} & {\color{red}{$2331.6_{\pm252}$}} & {\color{red}{$2891.5_{\pm357}$}} & {\color{red}{$934.7_{\pm20}$}} \\
\hline
\multirow{6}{*}{20} & Normal SAC          & $-83.1_{\pm9}$ & $-264.9_{\pm5}$ & $27.5_{\pm5}$ & $64.6_{\pm1}$ & $364.3_{\pm7}$ & $24.3_{\pm0}$ \\

                    & DC/AC             & $258.3_{\pm42}$ & $860.9_{\pm288}$ & $12.8_{\pm6}$ & $-2.9_{\pm5}$ & $237.3_{\pm73}$ & $4.1_{\pm0}$ \\

                    & Delayed-SAC  & $955.7_{\pm110}$ & $1377.8_{\pm140}$ & $1164.1_{\pm278}$ & $811.5_{\pm163}$ & $370.3_{\pm17}$ & {\color{blue}{$933.5_{\pm33}$}} \\
                    
                    & State Augmentation-MLP    &  $748.8_{\pm506}$ &  $915.1_{\pm101}$ &  $488.7_{\pm54}$ &  $442.2_{\pm191}$ & $455.7_{\pm6}$ &  $24.3_{\pm1}$ \\

                    & Conservative-BPQL ({\color{blue}ours}) & {\color{blue}{$1976.5_{\pm248}$}} & {\color{blue}{$3727.2_{\pm279}$}} & {\color{blue}{$1346.7_{\pm245}$}} & {\color{blue}{$1025.7_{\pm302}$}} & {\color{blue}{$1143.8_{\pm371}$}} & $566.9_{\pm88}$ \\      
                    
                    \cline{2-8}
                    & BPQL ({\color{red}constant}) & {\color{red}{$2078.9_{\pm157}$}} & {\color{red}{$3062.7_{\pm252}$}} & {\color{red}{$1526.7_{\pm227}$}} & {\color{red}{$846.7_{\pm443}$}} & {\color{red}{$1197.7_{\pm457}$}} & {\color{red}{$608.7_{\pm210}$}} \\                    
\Xhline{2\arrayrulewidth}
\end{tabular}}
\end{table*}

Table~\ref{result-table} presents the performance of baseline algorithms under uniformly distributed random delays with $\Delta_\text{max} = \{5, 10, 20\}$ on the continuous control tasks from the MuJoCo benchmarks. Empirical results reveal that the conservative-BPQL demonstrates the most remarkable performance across all tasks. In contrast, normal SAC, which does not account for the effect of delays, exhibits performance comparable to that of a random policy. Despite respectable performance in some tasks, delayed-SAC exhibits inconsistent and task-dependent performance. It may be attributed to the accumulation of prediction errors as task complexity or the degree of randomness in delay increases, underscoring the need for a more carefully designed dynamics model. Lastly, while DC/AC and state augmentation-MLP perform reasonably well, their effectiveness deteriorates substantially as the maximum delay increases. Additional results are given in \ref{additional-results}.

\begin{table*}[!h] 
\caption{The delay-free normalized scores.} \label{normalized-score}
\centering
\renewcommand{\arraystretch}{1.5}
\resizebox{\textwidth}{!}{
\begin{tabular}{ccccccccc}
\Xhline{3\arrayrulewidth}
\multicolumn{2}{c}{\textbf{Environment}} & \multirow{2}{*}{Ant-v3} &  \multirow{2}{*}{HalfCheetah-v3} & \multirow{2}{*}{Hopper-v3} & \multirow{2}{*}{Walker2d-v3} & \multirow{2}{*}{Humanoid-v3} & \multirow{2}{*}{InvertedPendulum-v2} & \multirow{2}{*}{\textbf{Average}}\\ 

\cline{1-2} 
$\Delta_\text{max}$ & \textbf{Algorithm}  & \\
\Xhline{1.5\arrayrulewidth}
\multirow{2}{*}{5} & BPQL ({\color{red}constant})  & $1.14$ & $0.62$ & $0.88$ & $0.74$ & $0.99$ & $1.00$ & {\color{red}0.90} \\
                    
                    & Conservative-BPQL ({\color{blue}random}) & $1.12$ & $0.68$ & $0.89$ & $0.85$ & $0.97$ & $0.99$ & {\color{blue}0.91}\\
                    
\Xhline{1.5\arrayrulewidth}
\multirow{2}{*}{10} & BPQL ({\color{red}constant})  & $0.86$ & $0.51$ & $0.88$ & $0.70$ & $0.89$ & $0.97$ & {\color{red}0.80} \\
                    
                    & Conservative-BPQL ({\color{blue}random})  & ${0.84}$ & $0.53$ & $0.94$ & $0.64$ & $0.87$ & $0.98$ & {\color{blue}0.80} \\           
                    
\Xhline{1.5\arrayrulewidth}
\multirow{2}{*}{20} & BPQL ({\color{red}constant})  & $0.64$ & $0.35$ & $0.63$ & $0.25$ & $0.34$ & $0.58$ & {\color{red}0.47} \\
                    
                    & Conservative-BPQL ({\color{blue}random}) & $0.61$ & $0.45$ & $0.55$ & $0.30$ & $0.32$ & $0.58$ & {\color{blue}0.48}\\                                      
\Xhline{3\arrayrulewidth}
\end{tabular}}
\end{table*}

Moreover, to support our claim in Proposition \ref{proposition}, we compare the conservative agent (conservative-BPQL) in random-delay MDPs with the normal agent (BPQL) in constant-delay MDPs with a fixed delay $\Delta = \Delta_{\max}$. For comparability, we report the delay-free normalized score, which is defined as
\begin{align}
    J_{\text{normalized}} = \frac{J_{\text{algorithm}}-J_{\text{random}}}{J_{\text{delay-free}}-J_{\text{random}}},
\end{align}where $J_{\text{algorithm}}$, $J_{\text{delay-free}}$, and $J_{\text{random}}$ represent the performance of each baseline algorithm, the delay-free SAC reference, and a random policy, respectively. The results in Table~\ref{normalized-score} show that the two agents achieved \textit{nearly identical} performance across all tasks. This provide strong empirical support for our argument that the conservative agent constructs a constant-delay surrogate under bounded random delays.

\subsection{Ablation Study} \label{Ablation}

\noindent\textbf{General Applicability} As aforementioned, any constant-delay method can be extended to random-delay MDPs with the aid of the conservative agent. To demonstrate this generality, we developed a second conservative RL algorithm, termed conservative-VDPO, which builds upon another state-of-the-art constant-delay method, called VDPO \cite{VDPO}. The corresponding results are reported in Table~\ref{VDPO-table}.

The results demonstrate that both the conservative RL algorithms achieve consistently strong empirical performance in random-delay MDPs. Given that these two algorithms (BPQL and VDPO) were not originally designed to handle random delays, these findings underscore the broad applicability of the conservative agent for extending conventional constant-delay methods to random-delay MDPs. Moreover, these results suggest substantial potential for further improvement as constant-delay methods continue to evolve.

\begin{table*}[!h] 
\caption{Results of two conservative RL algorithms evaluated for one million time steps with five random seeds.} \label{VDPO-table}
\centering
\renewcommand{\arraystretch}{1.5}
\resizebox{\textwidth}{!}{
\begin{tabular}{cccccccc}
\Xhline{3\arrayrulewidth}
\multicolumn{2}{c}{\textbf{Environment}} & \multirow{2}{*}{Ant-v3} &  \multirow{2}{*}{HalfCheetah-v3} & \multirow{2}{*}{Hopper-v3} & \multirow{2}{*}{Walker2d-v3} & \multirow{2}{*}{Humanoid-v3} & \multirow{2}{*}{InvertedPendulum-v2} \\ 

\cline{1-2} 
$\Delta_\text{max}$ & \textbf{Algorithm}  & \\

\Xhline{1.5\arrayrulewidth}
\multirow{2}{*}{5}                     & Conservative-BPQL ({\color{blue}ours}) & {{$3679.8_{\pm167}$}} &       {{$5583.9_{\pm169}$}} & $2174.1_{\pm155}$ & {{$2843.2_{\pm272}$}} & {{$3157.7_{\pm292}$}} & ${958.8}_{\pm14}$\\    

                    & Conservative-VDPO ({\color{blue}ours}) & {{$4373.3_{\pm181}$}} & {{$4871.9_{\pm41}$}} & $1942.1_{\pm115}$ & {{$3415.7_{\pm283}$}} & {{$2845.2_{\pm515}$}}  & ${774.2}_{\pm136}$ \\

\hline
\multirow{2}{*}{10} & Conservative-BPQL ({\color{blue}ours})  & {{$2744.5_{\pm112}$}} & {{$4511.8_{\pm143}$}} & {{$2300.9_{\pm164}$}} & {{$2122.3_{\pm292}$}} & {{$2820.5_{\pm348}$}} & $936.9_{\pm38}$ \\      
                    
                    & Conservative-VDPO ({\color{blue}ours}) & {{$3086.2_{\pm106}$}} & {{$3355.5_{\pm210}$}} & {{$1952.3_{\pm114}$}} & {{$2608.2_{\pm235}$}} & {{$2189.4_{\pm474}$}} & $610.9_{\pm114}$ \\    
\hline
\multirow{2}{*}{20} & Conservative-BPQL ({\color{blue}ours}) & {{$1976.5_{\pm248}$}} & {{$3727.2_{\pm279}$}} & {{$1346.7_{\pm245}$}} & {{$1025.7_{\pm302}$}} & {{$1143.8_{\pm371}$}} & $566.9_{\pm88}$ \\    
                    
                    & Conservative-VDPO ({\color{blue}ours}) & {{$2419.7_{\pm95}$}} & {{$2364.9_{\pm180}$}} & {{$1366.2_{\pm186}$}} & {{$804.8_{\pm129}$}} & {{$996.5_{\pm90}$}} & $19.4_{\pm8}$ \\                      
\Xhline{3\arrayrulewidth}
\end{tabular}}
\end{table*}

\medskip 

\noindent\textbf{Computational Overheads}
From an implementation perspective, the conservative agent differs from a normal agent only in that it requires a warm-up period of $\Delta_\text{max}$ steps at the beginning of each episode. The observed states are simply queued and each state is used exactly $\Delta_\text{max}$ time steps after its generation. This simple rule can be achieved by the priority queue, which supports operations with $O(\log N)$ time complexity, where $N$ denotes the number of stored states. Since the number of states in the queue is at most $\Delta_\text{max}$, we believe the computational overhead to be negligible in practice.

\begin{figure*}[!h] 
    \centering
    \includegraphics[width=0.95\textwidth]{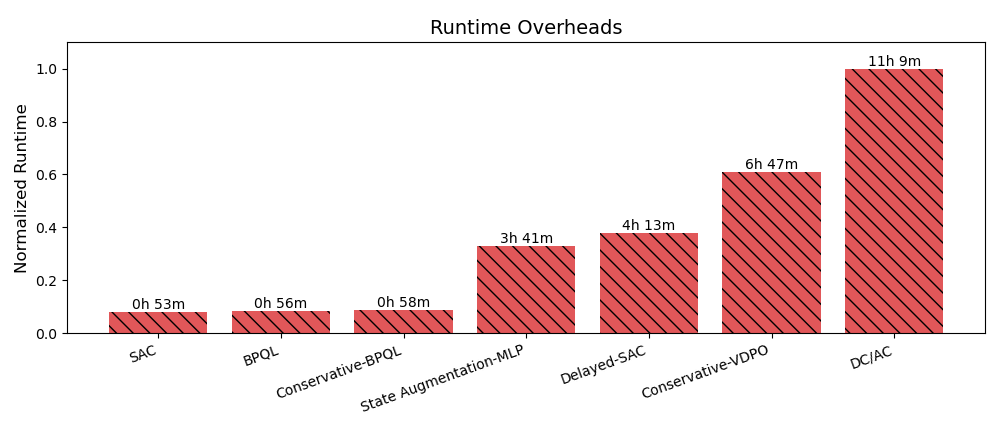}
    \caption{Runtime overheads of each algorithm measured over one million global time steps and averaged over five trials.} \label{runtimes-fig}
\end{figure*}

To quantify the computational overhead, we measured the runtime of baseline algorithms for five times on an NVIDIA RTX 3060 Ti GPU and an Intel i7-12700KF CPU. The results are reported in Fig. \ref{runtimes-fig}. Empirically, the two algorithms—BPQL and conservative-BPQL—exhibit no significant difference in runtime.

\section{Conclusion}
    
In this study, we investigated the challenge of reinforcement learning in environments with bounded random delays and proposed the conservative agent for robust decision-making under such conditions. We demonstrated that this agent can reformulate a random-delay MDP into its constant-delay surrogate, thereby providing a plug-and-play framework that allows any constant-delay methods to be directly extended to random-delay MDPs without algorithmic modification. Apart from a maximum delay, the conservative agent does not require any prior knowledge of the underlying delay distribution and maintains performance invariant to distributional changes as long as this maximum delay remains unchanged. This agent eliminates the need to estimate either individual delays or the delay distribution itself, thereby effectively addressing the difficulties induced by the inherent unpredictability of random delays. 

We presented a theoretical analysis of the conservative agent and derived a formal bound on the performance difference between the normal and conservative agents under bounded random delays. These results provide a guidance on when the conservative agent is well justified. We then investigated a limitation of our approach and introduced an effective mitigation strategy. The empirical results show that the proposed conservative RL algorithm achieved the strongest performance, consistently outperforming the state-of-the-art random-delay algorithms in terms of both asymptotic performance and sample efficiency. We believe our conservative agent can help researchers avoid treating constant-delay and random-delay environments as separate problems, thereby facilitating more focused and in-depth research in the field of delayed reinforcement learning.

\newpage 

\appendix 

\section{Discussion}

\subsection{Relation to the Freeze Agent} \label{freeze-agent}

Both the conservative and freeze approaches~\cite{augmented} share a common foundation in that they are built upon the classical complete information method and are designed to handle random delays. Nevertheless, these two approaches rely on different assumptions and employ conceptually different control mechanisms.

In the complete information method under a fixed delay $\Delta \in \mathbb{N}$, the state is augmented with the most recently observed state and the last $\Delta$ actions 
\begin{align}
    x_t = (s_{t-\Delta}, a_{t-\Delta}, a_{t-\Delta+1}, \dots, a_{t-1}), 
\end{align} 
so that the augmented state is Markovian. As mentioned before, the dimension of this augmented state either remains constant or increases by one at each time step under random delays, and thus can grow without bound in infinite-horizon problems if the delays are not assumed to be bounded. 

The freeze agent models delays as discrete, finite-valued random variables and assumes the existence of a maximum allowable dimension $N_\text{max}$ for the augmented state in order to keep its dimension bounded. Whenever the augmented state reaches this maximum allowable dimension, the freeze agent suspends decision-making and resumes it with a refreshed augmented state only when a timely state is observed. Therefore, the freeze approach can be viewed as a \textit{state-representation-level} control mechanism that keeps the dimension of the augmented state bounded. However, the resulting performance can be highly task-dependent and unstable, since the agent ignores the system's evolution during its inactive periods.

In contrast, the conservative agent assumes that delays are bounded by a known maximum delay $\Delta_\text{max}$. Apart from an initial warm-up period of $\Delta_\text{max}$ time steps at the beginning of each episode, it never freezes and continues to make decisions at every time step. The observed states are simply queued, and each state is used to form an augmented state exactly $\Delta_\text{max}$ time steps after its generation. Therefore, the conservative approach can be regarded as a \textit{policy-level} control mechanism that regularizes the timing of state usage. This approach ensures that decision-making remains dense in time and leads to more stable behavior than its freeze-based counterpart, since it does not discard information that would otherwise be ignored during frozen intervals in the freeze approach. Moreover, because it always constructs the same constant-delay surrogate, the performance of the conservative agent is invariant to changes in the delay distribution as long as the maximum delay remains unchanged. This delay-agnostic property further distinguishes the conservative agent from the freeze agent, whose behavior and performance can vary substantially with the underlying delay distribution through the frequency and timing of frozen periods.

\subsection{Additional Limitation} \label{limitation}

The conservative agent may experience learning inefficiencies under long-tailed delay distributions, as it does not utilize state information available at intermediate time steps. To mitigate this issue, a quantile-based cutoff strategy can be employed. In particular, if the delay follows a known long-tailed distribution such as a log-normal distribution, a pseudo maximum allowable delay $\Delta_\text{p.max}$ can be set at a reasonable upper quantile point, such that $\Delta_\text{p.max} \ll \Delta_\text{max}$. If a state is not observed by this threshold, the agent may instead reuse the most recently used state, assuming the underlying MDP exhibits sufficient smoothness. We anticipate that the potential performance degradation from this heuristic is acceptable relative to the significant sample inefficiency incurred by a large $\Delta_\text{max}$. Nonetheless, this strategy assumes prior knowledge of the delay distribution, which may not be practical in certain environments. Addressing this limitation remains an important direction for future research.

\newpage 

\section{Environmental and Implementation Details} \label{Experimental-detail} 

\begin{figure*}[h] 
\centering
\begin{subfigure}[t]{0.15\textwidth}
  \includegraphics[width=\linewidth]{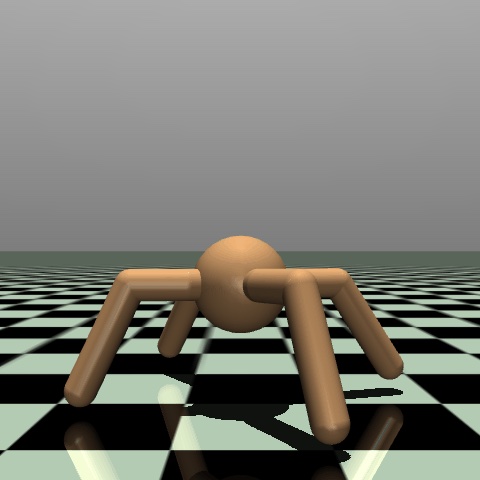}
  \caption{Ant-v3}
\end{subfigure}\hfill
\begin{subfigure}[t]{0.15\textwidth}
  \includegraphics[width=\linewidth]{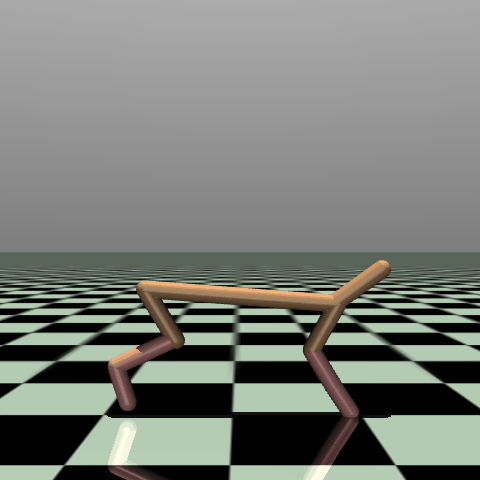}
  \caption{HalfCheetah-v3}
\end{subfigure}\hfill
\begin{subfigure}[t]{0.15\textwidth}
  \includegraphics[width=\linewidth]{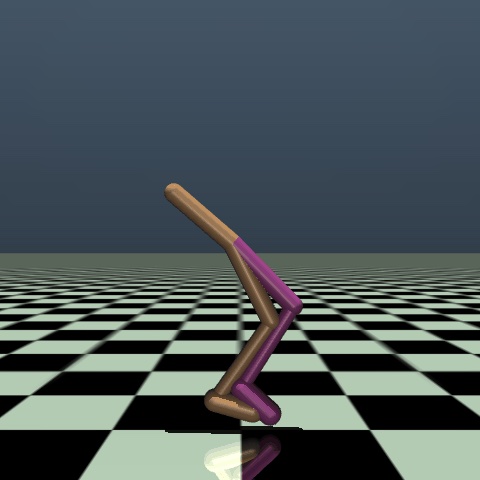}
  \caption{Walker2d-v3}
\end{subfigure}\hfill
\begin{subfigure}[t]{0.15\textwidth}
  \includegraphics[width=\linewidth]{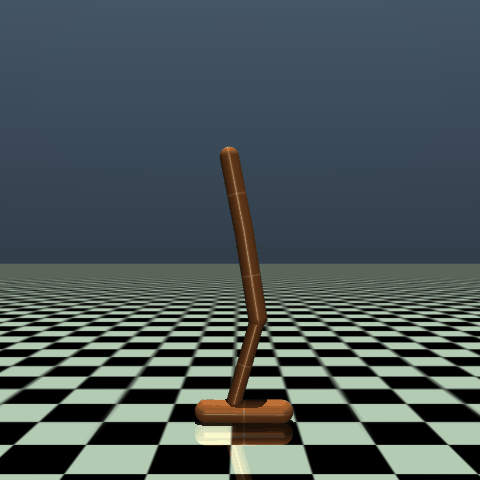}
  \caption{Hopper-v3}
\end{subfigure}\hfill
\begin{subfigure}[t]{0.15\textwidth}
  \includegraphics[width=\linewidth]{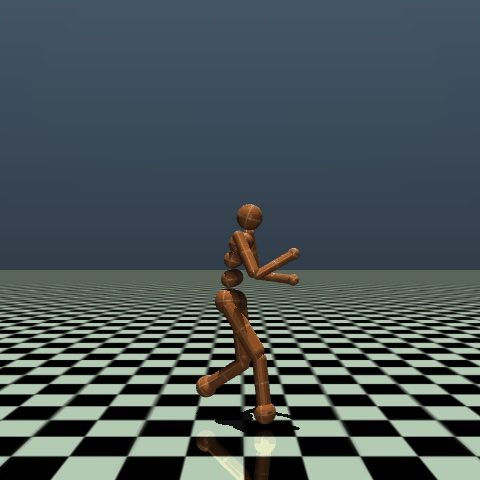}
  \caption{Humanoid-v3}
\end{subfigure}\hfill
\begin{subfigure}[t]{0.15\textwidth}
  \includegraphics[width=\linewidth]{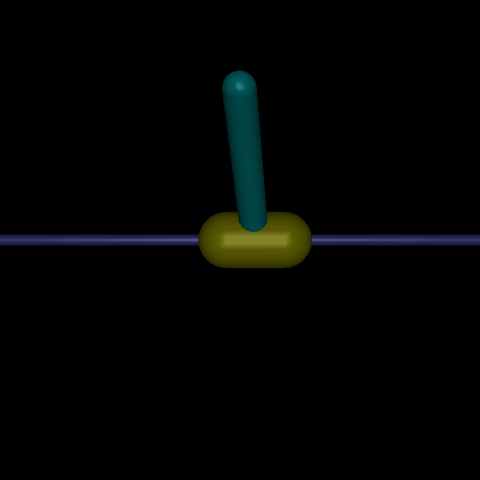}
  \caption{Pendulum-v2}
\end{subfigure}
\caption{Environments in the MuJoCo benchmark.} \label{environmental-2}
\end{figure*}

The implementation details of the conservative-BPQL align with those presented in \cite{BPQL}, with the specific hyperparameters listed in Table \ref{hyperparameter}. Since the random-delay baselines included in our experiments employ the SAC algorithm as their foundational learning algorithm, the hyperparameters are consistent across all approaches, except for the DC/AC and state augmentation-MLP algorithms, which were implemented with the source code provided by the authors. The environmental details are provided in Table \ref{environmental-1} and Fig. \ref{environmental-2}.

\begin{table}[!h]
\caption{Hyperparameters for the baseline algorithms.}\label{hyperparameter}
\centering
\renewcommand{\arraystretch}{1.1}
\begin{tabular}{cc}
\Xhline{3\arrayrulewidth}
\textbf{Hyperparameters} &  Values \\
\Xhline{1.5\arrayrulewidth}
Actor network  & 256, 256 \\
Critic network & 256, 256 \\
Learning rate (actor)  & 3e-4\\
Learning rate (beta-critic) & 3e-4\\
Temperature ($\alpha$)   & 0.2\\
Discount factor ($\gamma$)   & 0.99\\
Replay buffer size   & 1e6 \\
Mini-Batch size      & 256 \\
Target entropy       & -dim$|\mathcal{A}|$\\
Target smoothing coefficient ($\xi$) & 0.995\\
Optimizer            & Adam \cite{adam}\\
Total time steps     & 1e6 \\
\Xhline{3\arrayrulewidth}
\end{tabular}
\end{table}

\begin{table*}[!h] 
\caption{Environmental details of the MuJoCo benchmark.}\label{environmental-1}
\centering
\renewcommand{\arraystretch}{1.1}
\begin{tabular}{cccc}
\Xhline{3\arrayrulewidth}
\textbf{Environmet} &  State dimension & Action dimension & Time step (s)\\
\Xhline{1.5\arrayrulewidth}
Ant-v3              & 27 & 8 & 0.05\\ 
HalfCheetah-v3      & 17 & 6 & 0.05\\
Walker2d-v3         & 17 & 6 & 0.008\\
Hopper-v3           & 11 & 3 & 0.008\\
Humanoid-v3         & 376 & 17 & 0.015\\
InvertedPendulum-v2 & 4 & 1 & 0.04\\
\Xhline{3\arrayrulewidth}
\end{tabular}
\end{table*}

\newpage

\section{Pseudo-code of Conservative-BPQL} \label{Appendix-algorithm}

\begin{algorithm}[h!]
\caption{Conservative Belief Projection-based $Q$-Learning (Conservative-BPQL)} \label{algorithm}
\begin{algorithmic}
\State \textbf{Input:} actor $\pi_{\phi}(a \vert \tilde{x})$, beta-critic $Q_{\theta, \beta}(s, a)$, target beta critic $Q_{\tilde{\theta}, \beta}(s,a)$, replay buffer $\mathcal{D}$, temporary buffer $\mathcal{B}$, observation delay $\Delta = {\color{blue}\Delta_\text{max}}$, beta-critic learning rate $\psi_{Q}$, actor learning rate $\psi_{\pi}$, soft update rate $\xi$, episodic length $H$, and total number of episodes $E$.
\For {episode $e = 1$ to $E$}
    \For {time step $t = 1$ to $H$}
        \If{$t \leq \Delta$} 
        \Comment{{\color{blue}Wait for $\Delta$ time-steps}}
        \State select random or `no-ops' action $a_t$
        \State execute $a_t$ on environment
        \State put $a_t$, observed states, rewards to $\mathcal{B}$
        \Else
        \State get $s_{t-\Delta}, a_{t-\Delta}, \dots, a_{t-1}$ from $\mathcal{B}$
        \State $\tilde{x}_t \gets (s_{t-\Delta}, a_{t-\Delta}, \dots, a_{t-1})$ 
        \State $a_t \gets \pi_{\phi}(\tilde{x}_t)$
        \State execute $a_t$ on environment
        \State put $a_t$, observed states, rewards to $\mathcal{B}$
            \If{$t > 2\Delta$}
                \State get $s_{t-2\Delta}, s_{t-2\Delta+1}, s_{t-\Delta},r_{t-\Delta}, a_{t-2\Delta}, \dots, a_{t-\Delta}$ from $\mathcal{B}$
                \State $\tilde{x}_{t-\Delta} \gets (s_{t-2\Delta}, a_{t-2\Delta}, \dots, a_{t-\Delta})$ 
                \State $\tilde{x}_{t-\Delta+1} \gets (s_{t-2\Delta+1}, a_{t-2\Delta+1}, \dots, a_{t-\Delta+1})$ 
            \State store $(\tilde{x}_{t-\Delta}, s_{t-\Delta}, a_{t-\Delta}, r_{t-\Delta}, \tilde{x}_{t-\Delta+1}, s_{t-\Delta+1})$ in $\mathcal{D}$
            \State pop $s_{t-2\Delta}, a_{t-2\Delta}$ from $\mathcal{B}$
            \EndIf
        \EndIf
    \EndFor
    \For {each gradient step}
    \State $\theta \gets \theta - \psi_{Q} \nabla \mathcal{J}_{Q_{\beta}}(\theta)$ \Comment{Update beta-critic}
    \State $\phi \gets \phi - \psi_{\pi} \nabla \mathcal{J}_{\pi}(\phi)$ \Comment{Update actor}
    \State $\tilde{\theta} \gets \xi\theta + (1-\xi)\tilde{\theta}$ \Comment{Update target beta-critic}
    \EndFor    
\EndFor    
\State \textbf{Output:} actor $\pi_{\phi}$
\end{algorithmic}
\end{algorithm}

The conservative agent can be combined with the BPQL framework with minimal modifications. Following \cite{BPQL}, a temporary buffer $\mathcal{B}$ has been employed to store the observed states, corresponding rewards, and historical actions, which enables the agent to access timely and delay-relevant historical information for constructing augmented states. Notably, the augmented reward can be empirically estimated through the use of replay buffer $\mathcal{D}$
\begin{align}
    \mathcal{R}_{\Delta}(\tilde{x}_{t-{\Delta}}, a_{t-{\Delta}}) \approx \mathbb{E}_{\mathcal{D}}[\mathcal{R}(s_{t-{\Delta}}, a_{t-{\Delta}})].
\end{align}Consequently, training the beta-critic and actor requires only the following set of experience tuples: 
\begin{align}
(\tilde{x}_{t-\Delta}, s_{t-\Delta}, a_{t-\Delta}, r_{t-\Delta}, \tilde{x}_{t-\Delta+1}, s_{t-\Delta+1}).    
\end{align}

\newpage 

\section{Performance Curves} \label{additional-results}

In this section, we represent the performance curves of each algorithm evaluated on the MuJoCo tasks under random delays with $\Delta_{\text{max}} \in \{5, 10, 20\}$. All tasks were conducted with five different seeds for one million time steps, where the shaded regions represent the standard deviation of average returns. 

\begin{figure}[!h]
    \centering
    \includegraphics[width=0.915\textwidth]{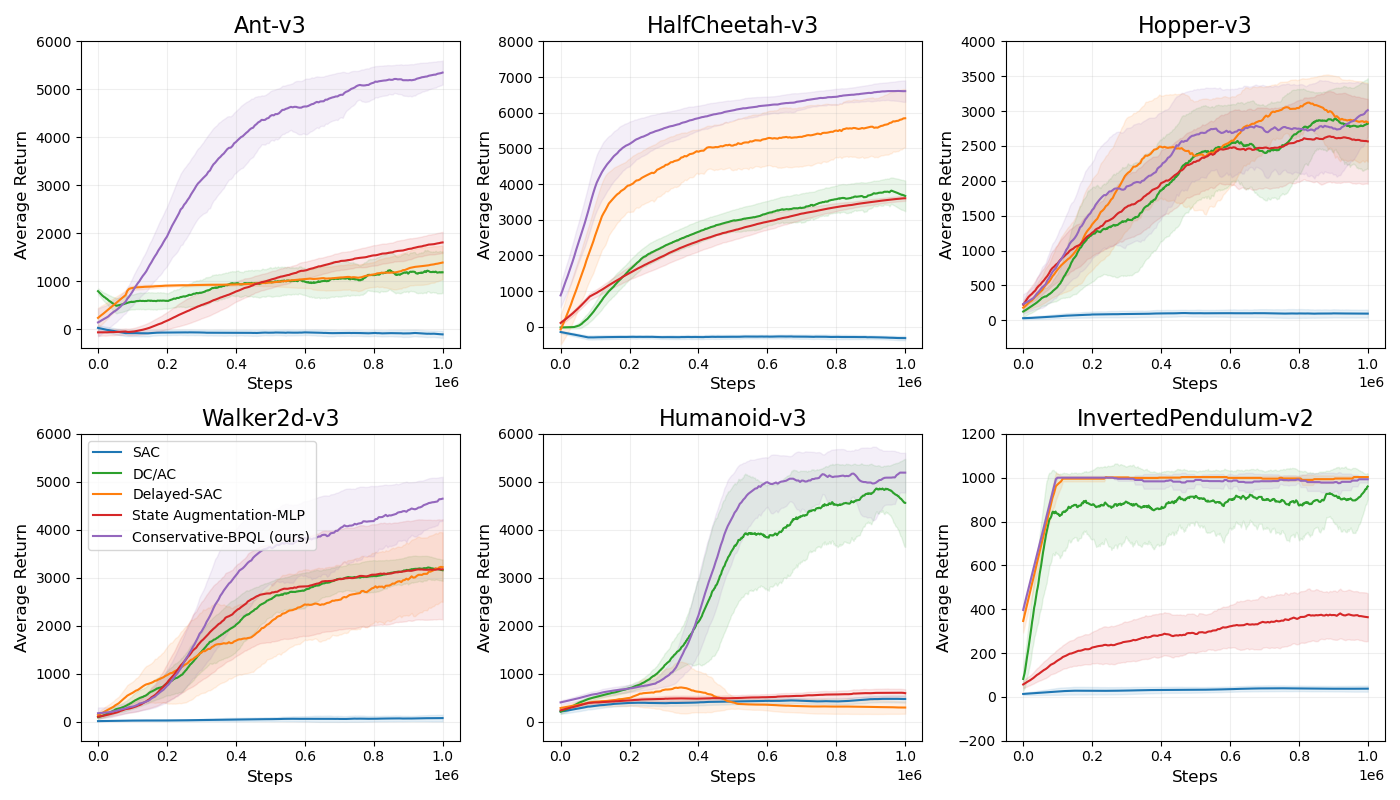} 
\caption{Performance curves of each algorithm on the MuJoCo tasks with $\Delta_{\text{max}} = 5$.}
\end{figure}

\begin{figure}[!h]
    \centering
    \includegraphics[width=0.915\textwidth]{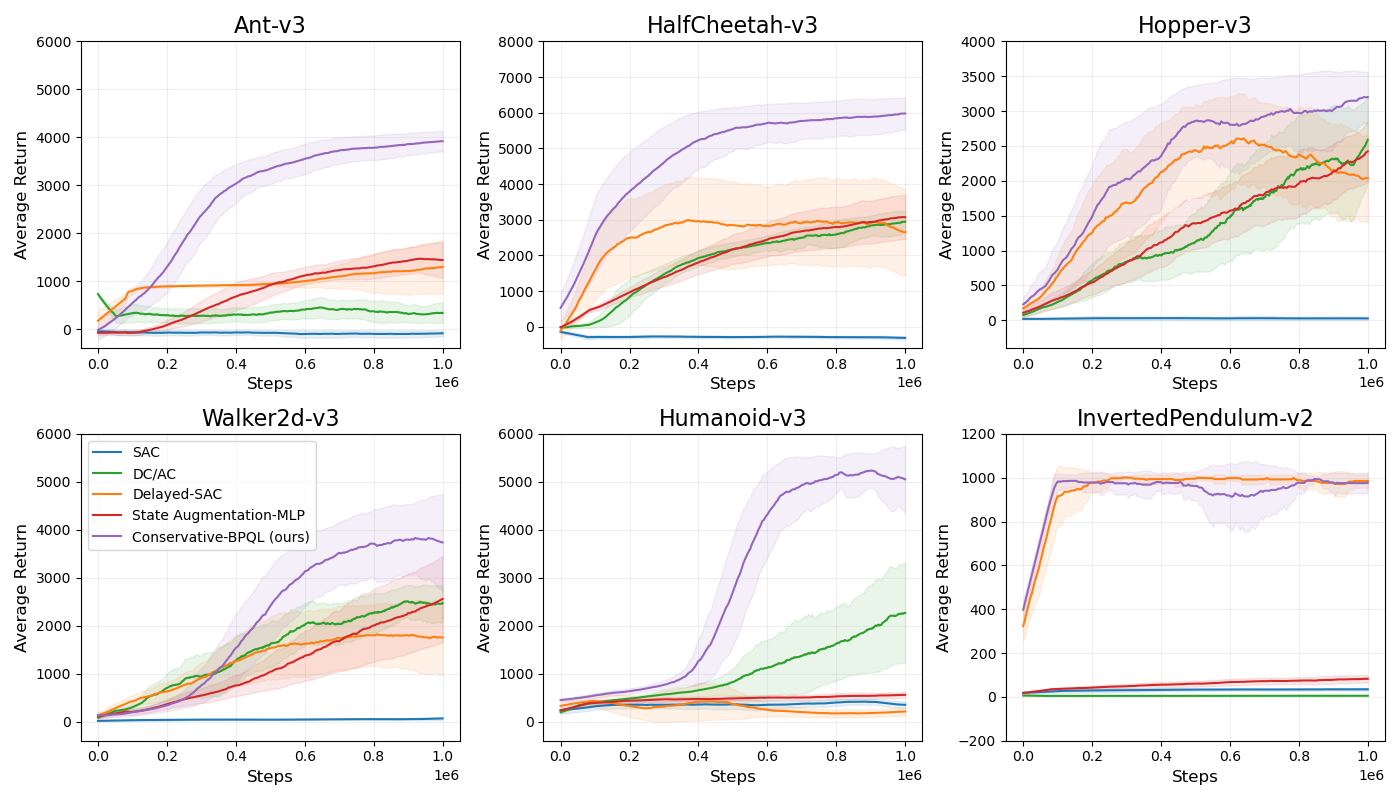}
\caption{Performance curves of each algorithm on the MuJoCo tasks with $\Delta_{\text{max}} = 10$.}
\end{figure}

\begin{figure}[!h]
    \centering
    \includegraphics[width=0.915\textwidth]{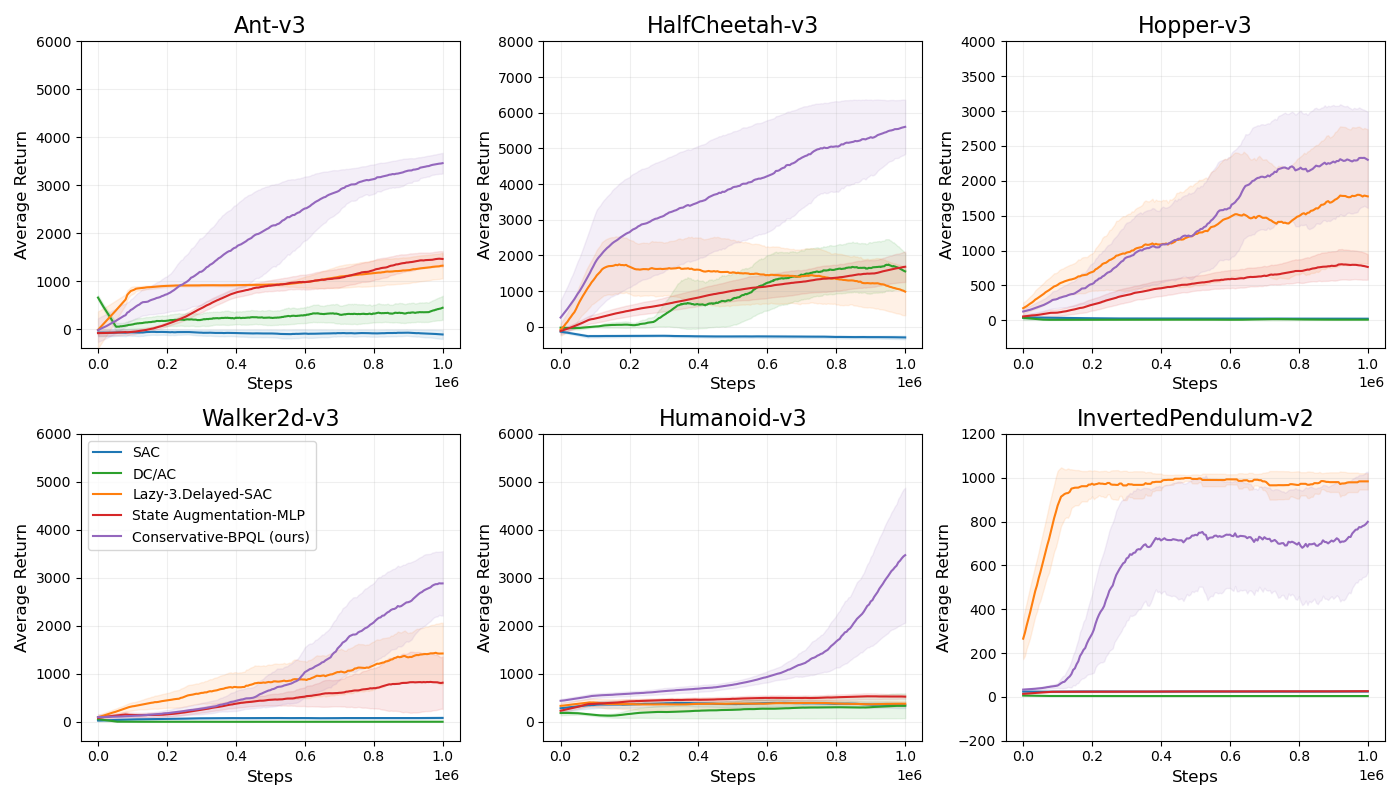} 
\caption{Performance curves of each algorithm on the MuJoCo tasks with $\Delta_{\text{max}} = 20$. }
\end{figure}

\newpage 

\section{Proofs}  \label{proof}

\subsection{Proof for Proposition 3.1} \label{proof-proposition-1}

\noindent\textbf{Proposition 3.1.} \textit{Let $\lambda$ be a delay distribution supported on the set $\Lambda = \{1, 2, \dots, \Delta_\text{max}\}$. If the agent follows the conservative decision-making strategy that assumes $\tau(s_{n}) = n+\Delta_{\text{max}}$ for all $n > 0$, then a random-delay MDP can be reformulated as a constant-delay surrogate with fixed delay $\Delta = \Delta_\text{max}$, which can be further reduced to an equivalent delay-free MDP.}

\medskip 

\begin{proof}[Proof sketch]
Suppose the augmented state at time $t = n + \Delta_\text{max}$ is given by
\begin{align} \label{eq:new-augmented-state-0}
\tilde{x}_{t} = (s_{n}, a_{n}, a_{n+1}, \dots, a_{t-1}).
\end{align} Under the conservative assumption, we have
\begin{align}
  \mathbb{P}(\Delta^{(n+1)}\leq\tau(s_n)-n) = \mathbb{P}(\Delta^{(n+1)}\leq \Delta_\text{max}) = 1,
\end{align} which determines the subsequent augmented state as
\begin{align}
\tilde{x}_{t+1} = (s_{n+1}, a_{n+1}, a_{n+2}, \dots, a_{t}).
\end{align} By induction, the same shift structure holds at every time step. Substituting $n = t-\Delta_\text{max}$ yields
\begin{align} 
\tilde{x}_{t} &= (s_{t-\Delta_\text{max}}, a_{t-\Delta_\text{max}}, a_{t-\Delta_\text{max}+1}, ..., a_{t-1}), \quad \forall t>\Delta_\text{max}. 
\end{align} This formulation matches the definition of augmented states in constant-delay MDP with $\Delta = \Delta_\text{max}$, which can be further reduced to equivalent delay-free MDPs. 
\end{proof}

\subsection{Proof for Theorem 3.3} \label{proof-theorem-1}

\noindent\textbf{Theorem 3.3.} \textit{Let $\gamma\in(0,1)$ and $r_t\in[0,R_{\max}]$ for all $t\ge 0$. Then for any $\Delta, \Delta' \in \Lambda$,
\begin{align}
    \left|J^*_{\Delta} - J^*_{\Delta'}\right|  \leq  \frac{\gamma^{\min(\Delta, \Delta')}}{1 - \gamma}R_\text{max}\left|\Delta-\Delta'\right|.
\end{align} Moreover, it extends to
\begin{align}
    0 \leq J_\text{meta} - J^*_\text{max} \leq \frac{\gamma^{\Delta_\text{min}}}{1 - \gamma}R_\text{max}\left(\Delta_\text{max} - \mathbb{E}_{\lambda}\left[\Delta\right]\right), \label{bound-1}
\end{align} where $\Delta_\text{min}$ and $\Delta_\text{max}$ denote the minimum and maximum supports of the delay distribution $\lambda$.} 

\medskip 

\begin{proof}
Let $\mathcal{M}_\Delta = (\mathcal{X}_\Delta, \mathcal{A}, \mathcal{P}_\Delta, \mathcal{R}_\Delta, \gamma)$ be a delay-free reformulation of constant-delay MDP $\mathcal{M}^+ = (\mathcal{M}, \Delta)$, $\mathcal{M} = (\mathcal{S}, \mathcal{A}, \mathcal{P}, \mathcal{R}, \gamma)$ be the underlying delay-free MDP with bounded reward $r_t \in [0, R_{\text{max}}]$ for all $t \geq 0$, and $\mathcal{M}_{\lambda} = (\mathcal{M}, \lambda, \tau)$ be the random-delay MDP with delay distribution $\lambda$ supported on $\Lambda = \{1, 2, \dots, \Delta_\text{max}\}$. For any optimal policy $\pi^*_\Delta$ in $\mathcal{M}_\Delta$, we define a one-step-lagged policy $L(\pi^*_\Delta)$ working in $\mathcal{M}_{\Delta+1}$ by
\begin{align}
    L\left(\pi^*_\Delta\right)\left(x^{(\Delta+1)}_t\right) \; \triangleq \; \pi^*_\Delta\left(f_L\left(x^{(\Delta+1)}_t\right)\right) \; = \; \pi^*_\Delta\left(x^{(\Delta)}_{t-1}\right)
\end{align} for all $ t \geq 1$, where $x^{(\Delta+1)}_t \in \mathcal{X}_{\Delta+1}$ is the augmented state defined in $\mathcal{M}_{\Delta+1}$ and a mapping $f_L: \mathcal{X}_{\Delta+1} \rightarrow \mathcal{X}_{\Delta}$ removes the most recent action from the augmented state. For example, 
\begin{align}
 f_L\left(x^{(\Delta+1)}_t\right) &= f_L\left(s_{t-\Delta-1},a_{t-\Delta-1},a_{t-\Delta},\dots,a_{t-2},a_{t-1}\right)\\
 &= \left(s_{t-\Delta-1},a_{t-\Delta-1}, a_{t-\Delta},\dots, a_{t-2}\right)\\ 
 &= x^{(\Delta)}_{t-1} \in\mathcal{X}_\Delta.
\end{align} Intuitively, $L(\pi^*_\Delta)$ imitates the decisions that $\pi^*_\Delta$ would have taken one step earlier under a fixed delay $\Delta$. We use this policy to upper-bound the performance loss per unit increase in delay. Firstly, we have
\begin{align}
    \left\vert r^{(\Delta +1)}_t - r^{(\Delta) }_{t} \right\vert \; \leq \; R_\text{max}, \quad \forall  t \geq 0,
\end{align} where $r^{(\Delta)}_t \in \mathcal{R}_{\Delta}$ and $r^{(\Delta+1)}_t \in \mathcal{R}_{\Delta+1}$. Under a cold-start protocol, in which the agent receives zero rewards until the augmented state becomes well defined, the performance gap is given by
\begin{align}
    \left\vert J^{*}_\Delta - J^{L}_{\Delta+1} \right\vert  &=  \left\vert \mathbb{E}_{\omega}\left[\sum^{\infty}_{t = \Delta}\gamma^{t}r^{(\Delta)}_t - \sum^{\infty}_{t = \Delta + 1}\gamma^{t}r^{(\Delta+1)}_t\right]\right\vert  \\
    &=  \left\vert \mathbb{E}_{\omega}\left[\gamma^{\Delta}r^{(\Delta)}_\Delta + \sum^{\infty}_{t = \Delta + 1}\gamma^{t}\left(r^{(\Delta)}_t - r^{(\Delta+1)}_t\right)\right]\right\vert \\
    & \leq  \mathbb{E}_{\omega}\left[\gamma^{\Delta}r^{(\Delta)}_\Delta + \sum^{\infty}_{t = \Delta + 1}\gamma^{t}\left\vert r^{(\Delta)}_t - r^{(\Delta+1)}_t\right\vert \right]\\
    & \leq  \mathbb{E}_{\omega}\left[\gamma^{\Delta}R_\text{max} + \sum^{\infty}_{t = \Delta + 1}\gamma^{t}R_\text{max}\right] \\
    & = \frac{\gamma^{\Delta}}{1 - \gamma} R_\text{max}.
\end{align} where $J^{L}_{\Delta+1}$ denotes the performance of $L(\pi^*_\Delta)$. Note that we couple the processes on a common probability space, where $\omega$ represents the fixed environmental stochasticity such as initial state and transitions, which allows for a consistent term-wise comparison. Since $J^*_{\Delta+1} \geq J^{L}_{\Delta+1}$, we have 
\begin{align}
    J^{*}_\Delta - J^*_{\Delta+1}  \leq J^{*}_\Delta - J^{L}_{\Delta+1} \leq \frac{\gamma^{\Delta}}{1 - \gamma}R_\text{max}.
\end{align} By symmetry, we obtain the two-sided bound
\begin{align}
    \left\vert J^{*}_\Delta - J^*_{\Delta+1} \right\vert \leq \frac{\gamma^{\Delta}}{1 - \gamma}R_\text{max},
\end{align} which yields an upper-bound on the performance degradation incurred by increasing the delay by one. Then for any $\Delta, \Delta' \in \Lambda$, a telescoping sum gives the result 
\begin{align}
    \left\vert J^*_\Delta - J^*_{\Delta'} \right\vert \; \leq \;\sum^{\vert\Delta - \Delta'\vert-1}_{i = 0} \left\vert J^*_{\min(\Delta, \Delta')  + i} - J^*_{\min(\Delta, \Delta') + i + 1} \right\vert \; \leq \; \frac{\gamma^{\min(\Delta, \Delta')}}{1 - \gamma}R_\text{max} \left\vert \Delta - \Delta'\right\vert. \label{result-1}
\end{align} As a result, we obtain the following bound 
\begin{align}
    J^*_{\Delta} - J^*_{\text{max}} \; \leq \; \frac{\gamma^{\Delta}}{1 - \gamma}R_\text{max}(\Delta_\text{max} - \Delta), \quad (\because \Delta \leq \Delta_\text{max}).
\end{align} Taking the expectation over the delay distribution $\lambda$
\begin{align}
     J_{\text{meta}} - J^*_{\text{max}} = \mathbb{E}_\lambda\left[J^*_\Delta - J^*_\text{max}\right] &\leq  \mathbb{E}_{\lambda}\left[\frac{\gamma^{\Delta}}{1 - \gamma}R_\text{max}(\Delta_\text{max} - \Delta)\right] \\ 
    &\leq  \frac{\gamma^{\Delta_\text{min}}}{1 - \gamma}R_\text{max}\cdot\mathbb{E}_{\lambda}\left[(\Delta_\text{max} - \Delta)\right] \\
    &=  \frac{\gamma^{\Delta_\text{min}}}{1 - \gamma}R_\text{max}\left(\Delta_\text{max} - \mathbb{E}_{\lambda}[\Delta]\right),
\end{align} where $\Delta_\text{min}$ denotes a minimum support of $\lambda$. Therefore, we have
\begin{align}
    0 \leq J_{\text{meta}} - J^*_{\text{max}} \leq \frac{\gamma^{\Delta_\text{min}}}{1 - \gamma}R_\text{max}\left(\Delta_\text{max} - \mathbb{E}_{\lambda}[\Delta]\right),
\end{align} where the left inequality holds by Lemma~\ref{theorem-1}. This completes the proof.    
\end{proof}

\subsection{Proof for Theorem 3.4} \label{proof-theorem-2}

\noindent\textbf{Theorem 3.4.} \textit{Let $\gamma\in(0,1)$ and $r_t\in[0,R_{\max}]$ for all $t\ge 0$. Then,
\begin{align}
0 \leq J_\text{meta} - J^*_{\text{max}}\leq\frac{R_\text{max}}{1 - \gamma}(1 - \gamma^{\Delta_\text{max}}),
\end{align} where $\Delta_\text{max}$ denotes the maximum support of the delay distribution $\lambda$.}

\medskip 

\begin{proof}
Let $\mathcal{M}_\Delta = (\mathcal{X}_\Delta, \mathcal{A}, \mathcal{P}_\Delta, \mathcal{R}_\Delta, \gamma)$ be a delay-free reformulation of constant-delay MDP $\mathcal{M}^+ = (\mathcal{M}, \Delta)$, $\mathcal{M} = (\mathcal{S}, \mathcal{A}, \mathcal{P}, \mathcal{R}, \gamma)$ be the underlying delay-free MDP with bounded reward $r_t \in [0, R_{\text{max}}]$ for all $t \geq 0$, and $\mathcal{M}_{\lambda} = (\mathcal{M}, \lambda, \tau)$ be the random-delay MDP with delay distribution $\lambda$ supported on $\Lambda = \{1, 2, \dots, \Delta_\text{max}\}$. For any optimal policy $\pi^*$ in $\mathcal{M}$, the state-value function under a horizon $H > \Delta_\text{max}$ is upper-bounded by
\begin{align}
V^{\pi^*}(s) = \mathbb{E}\left[\sum^{H-1}_{k = 0}\gamma^k r_{t+k} \mid s_t = s, \pi^* \right] \leq \sum^{H-1}_{k = 0}\gamma^k R_\text{max}, \quad \forall s \in \mathcal{S}. \label{eq:episodic-v-upper}
\end{align}
Under a cold-start protocol, the state-value function for an optimal policy $\pi_\Delta^*$ in $\mathcal{M}_\Delta$ is upper-bounded by
\begin{align}
    V^{\pi^*_{\Delta}}(x) \leq \sum^{H-1}_{k = \Delta}\gamma^{k}R_\text{max}, \quad \forall x \in \mathcal{X}_\Delta.
\end{align} Taking the expectation over the initial state distribution $\rho_\Delta$ induced by $\pi^*_\Delta$, we have
\begin{align}
J^*_\Delta  = \mathbb{E}_{\rho_{\Delta}}\left[V^{\pi^*_{\Delta}}(x)\right]\leq  \sum_{k=\Delta}^{H-1} \gamma^{k}R_{\text{max}}.
\end{align} Then for two constant delays $\Delta$ and $\Delta_\text{max}$, 
\begin{align}
J^*_\Delta - J^*_\text{max} \leq R_{\max} \left( \sum_{k=\Delta}^{H-1} \gamma^k - \sum_{k=\Delta_{\max}}^{H-1} \gamma^k \right) &= R_{\text{max}}\sum_{k=\Delta}^{\Delta_{\text{max}} - 1} \gamma^{k}. \label{geo-sum}
\end{align} Since it satisfies
\begin{align}
R_{\text{max}}\sum_{k=\Delta}^{\Delta_{\text{max}} - 1} \gamma^{k} \leq R_{\text{max}}\sum_{k=0}^{\Delta_\text{max}-1} \gamma^{k} = R_{\text{max}}\frac{1 - \gamma^{\Delta_\text{max}}}{1 - \gamma}.
\end{align} Taking the expectation over the delay distribution $\lambda$
\begin{align}
J_\text{meta} - J^*_{\text{max}} \leq \frac{R_{\text{max}}}{1 - \gamma}(1 - \gamma^{\Delta_\text{max}}).
\end{align} Since $J_\text{meta} \geq J^*_{\text{max}}$ by Lemma~\ref{theorem-1}, we obtain
\begin{align}
0~\leq~J_\text{meta} - J^*_{\text{max}}~\leq~\frac{R_\text{max}}{1 - \gamma}(1 - \gamma^{\Delta_\text{max}}),
\end{align}which completes the proof.
\end{proof}

\newpage

\bibliographystyle{IEEEtran}
\bibliography{Bibliography/CA}

\end{document}